\documentclass[preprint,a4paper,fleqn]{cas-dc}
\ExplSyntaxOn
\cs_gset:Npn \__first_footerline: { 
  \group_begin: 
  \small \sffamily \__short_authors: 
  \group_end: 
}
\ExplSyntaxOff

\usepackage[authoryear]{natbib}

\usepackage{algorithm}
\usepackage{algpseudocode}
\usepackage{amsmath}
\usepackage{amssymb}
\usepackage{graphicx}
\usepackage{subcaption} 

\def\tsc#1{\csdef{#1}{\textsc{\lowercase{#1}}\xspace}}
\tsc{WGM}
\tsc{QE}

\begin{document}
\let\WriteBookmarks\relax
\def\floatpagepagefraction{1}
\def\textpagefraction{.001}

\shorttitle{}

\shortauthors{}  

\title [mode = title]{OpenMAP-HIE: Pathology Informed Score-based Diffusion Model for Unsupervised Neonatal Hypoxic-Ischemic Encephalopathy Lesion Segmentation}

\author[1]{Pengyu Kan}[orcid=0009-0009-0774-2588]
\credit{Conceptualization, Formal analysis, Investigation, Methodology, Software, Visualization, Writing - Original Draft, Writing – review and editing}

\affiliation[1]{organization={Department of Computer Science},
            addressline={Johns Hopkins University}, 
            city={Baltimore},
            citysep={}, 
            postcode={}, 
            state={MD},
            country={United States}}

\author[1,2]{Craig Jones}[orcid=0000-0002-0629-3006]
\credit{Conceptualization, Formal analysis, Methodology, Writing – review and editing}





\affiliation[2]{organization={Department of Biomedical Engineering},
            addressline={University of Alberta}, 
            city={Edmonton},
          citysep={}, 
            postcode={}, 
            state={AB},
            country={Canada}}

\author[3,4]{Kenichi Oishi}[orcid=0000-0002-1200-8611]
\credit{Conceptualization,  Formal analysis, Methodology, Funding acquisition, Project administration, Supervision, Writing – review and editing}


\ead{koishi2@jhmi.edu}



\affiliation[3]{organization={Department of Radiology},
            addressline={Johns Hopkins University School of Medicine}, 
            city={Baltimore},
          citysep={}, 
            postcode={}, 
            state={MD},
            country={United States}}
\affiliation[4]{organization={Department of Neurology},
            addressline={Johns Hopkins University School of Medicine}, 
            city={Baltimore},
          citysep={}, 
            postcode={}, 
            state={MD},
            country={United States}}

\cormark[1]

\cortext[1]{Corresponding author}



\begin{abstract}
Accurate localization of neonatal hypoxic--ischemic encephalopathy (HIE) lesions on apparent diffusion coefficient (ADC) images is clinically important but is limited by scarce expert annotations and substantial variation in lesion appearance. We propose OpenMAP-HIE, a pathology-informed score-based diffusion model for unsupervised HIE lesion localization without voxel-wise lesion supervision. A denoising diffusion model learns the healthy neonatal ADC distribution, while a ControlNet-based conditional branch preserves patient-specific anatomy. At inference, anomaly evidence is derived directly from differences between conditional and unconditioned noise predictions at independently evaluated diffusion noise levels, avoiding dependence on a final pseudo-healthy reconstruction or sequential reverse-diffusion chain. Based on the characteristic ADC reduction associated with acute HIE, only discrepancies consistent with the expected pathological direction are retained and aggregated across noise levels using a geometric mean. Spatial mean filtering further suppresses isolated responses. The framework was evaluated on the BONBID-HIE dataset. The ADC-only method achieved a Dice score of $0.283\pm0.244$, average precision of $0.505$, Hausdorff distance of $25.45\pm15.63$~mm, mean average surface distance of $9.40\pm8.49$~mm, and normalized surface distance of $0.355\pm0.236$, providing the best overlap and surface-distance performance among the evaluated unsupervised methods and Dice performance comparable to supervised ADC-only models. Optional population-derived ADC $Z$-map refinement further increased Dice to $0.602\pm0.269$. These findings support pathology-informed score discrepancies as a useful signal for neonatal HIE lesion localization, while highlighting remaining challenges in detecting small and low-contrast lesions.

\end{abstract}

\begin{keywords}
Hypoxic--ischemic encephalopathy \sep Neonatal brain MRI \sep Unsupervised anomaly detection \sep Diffusion models \sep Pathology informed modeling \sep Score-based modeling
\end{keywords}

\maketitle

\section{Introduction}\label{sec:intro}
\begin{figure*}
    \centering
    \includegraphics[width=0.8\linewidth]{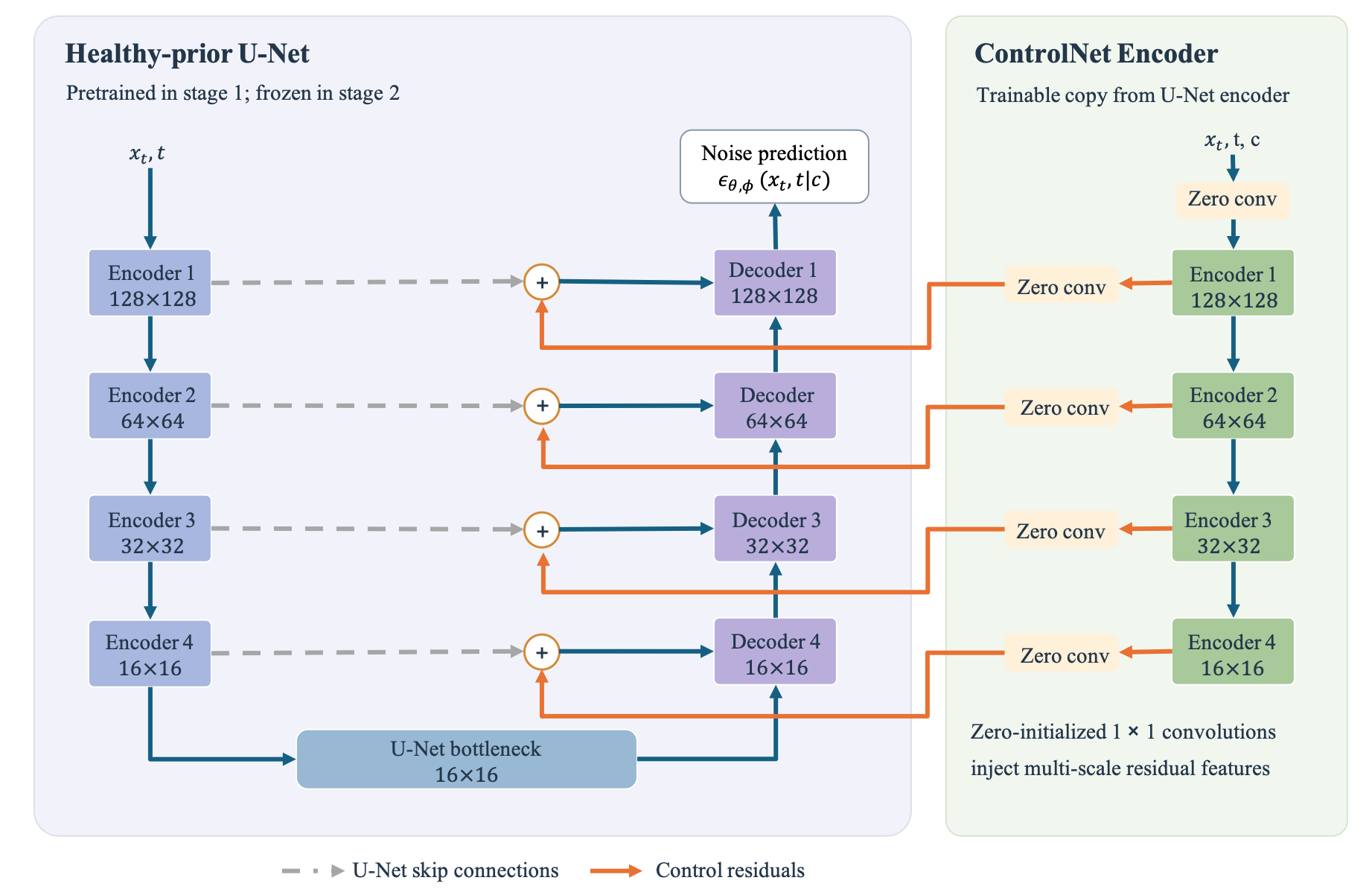}
    \caption{Architecture of the anatomy-conditioned diffusion model.
The healthy-prior U-Net is pretrained in stage~1 and frozen during
stage~2. A trainable ControlNet encoder, initialized from the U-Net
encoder, incorporates the spatial condition $\mathbf{c}$ alongside
the noisy input $\mathbf{x}_t$ and timestep $t$.
Multi-scale ControlNet features are passed through zero-initialized
$1\times1$ convolutions and added to the U-Net skip features,
enabling subject-specific conditioning of the noise prediction.
Dashed gray lines denote U-Net skip connections, and orange arrows
indicate ControlNet residual connections.}
    \label{fig:architecture}
\end{figure*}

\begin{figure}
    \centering
    \begin{subfigure}[b]{0.52\textwidth}
        \centering
        \includegraphics[width=1.0\linewidth]{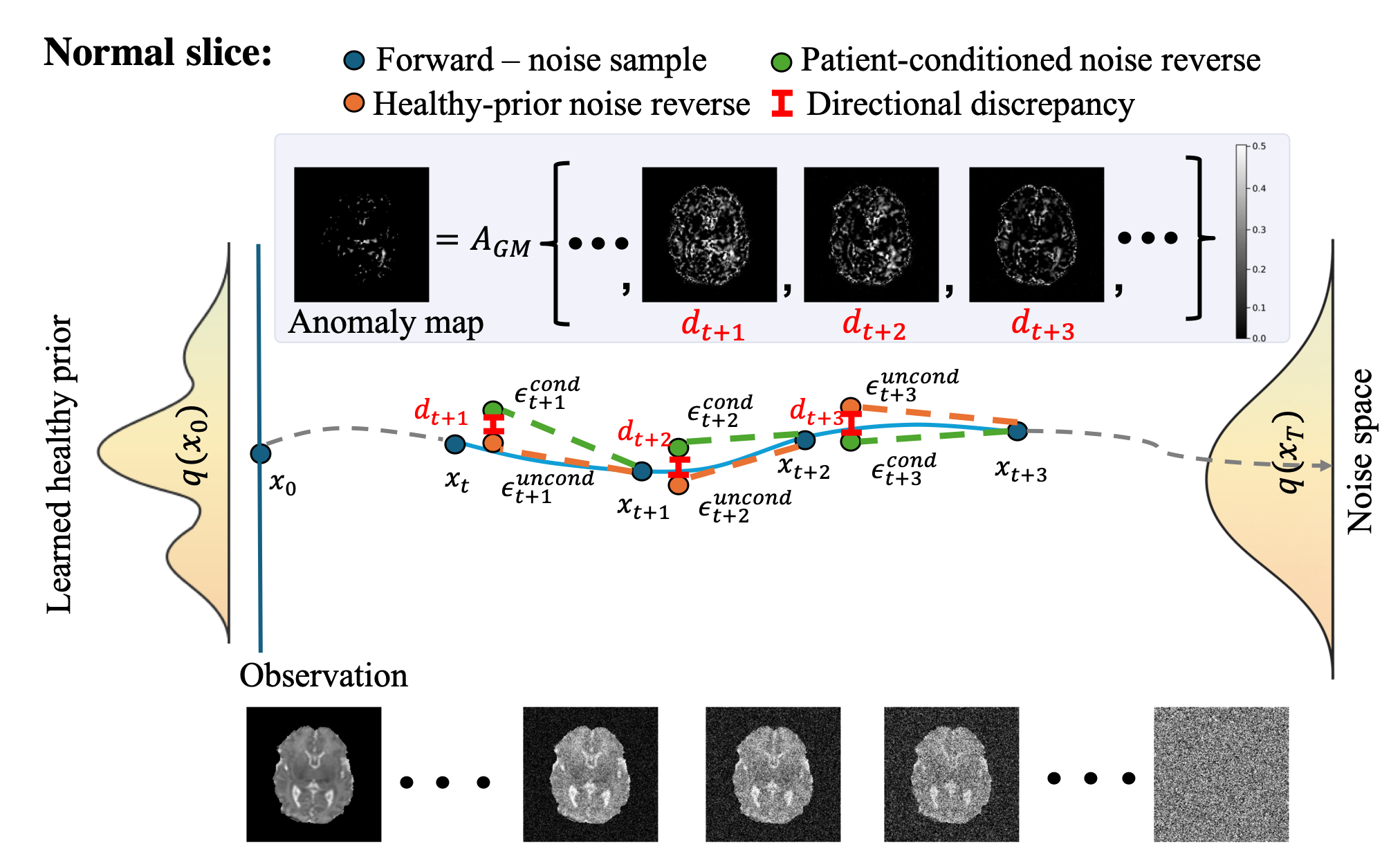}
        \caption{}
        \label{fig:normal_discrepancy}
    \end{subfigure}
    \hfill 
    \begin{subfigure}[b]{0.52\textwidth}
        \centering
        \includegraphics[width=1.0\linewidth]{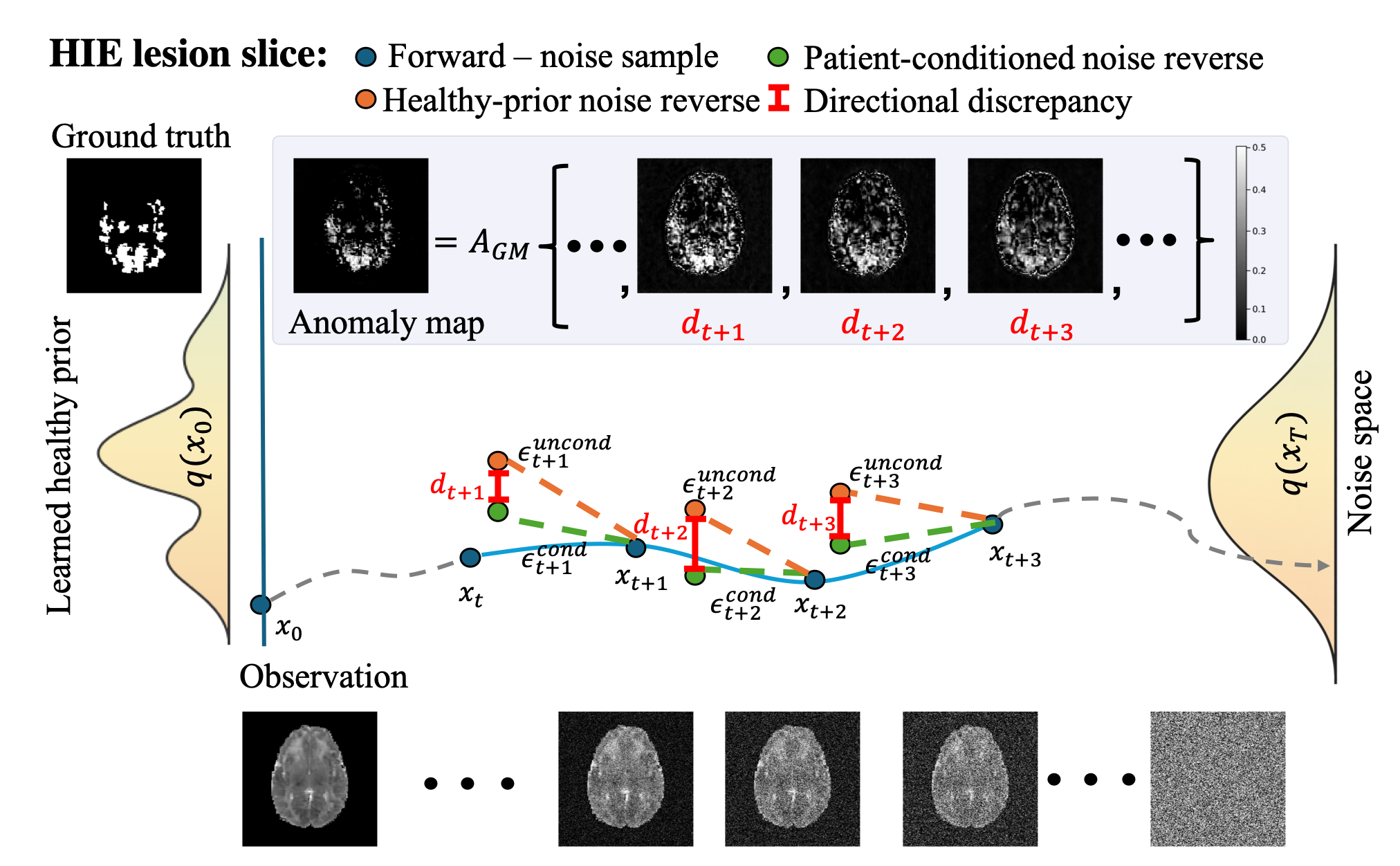}
        \caption{}
        \label{fig:lesion_discrepancy}
    \end{subfigure}
    
    \caption{
Illustration of inference using pathology-informed noise-prediction discrepancy aggregation for (a) a normal slice and (b) a slice containing HIE lesions.
The observed ADC image $\mathbf{x}_0$ is independently perturbed at
multiple noise levels. At each level, the patient-conditioned (green dashed line) and
healthy-prior (red dashed line) branches evaluate the same noisy input, and their
predicted noise difference is retained only in the direction consistent with the reduced-ADC prior.
Normal tissue is expected to exhibit relatively small discrepancies,
whereas lesion regions may produce larger, directionally consistent
responses across noise levels.
The resulting discrepancy maps are aggregated using the geometric
mean, $A_{\mathrm{GM}}$, to obtain the anomaly map.
Blue points denote noisy inputs, green and orange markers denote
patient-conditioned and healthy-prior predictions, respectively,
and red brackets indicate directional discrepancies.
The curves provide a sequential
forward-diffusion trajectory. The lesion ground truth annotation in (b) is shown
for reference only.
}
    \label{fig:discrepancy_concept}
\end{figure}

Hypoxic ischemic encephalopathy (HIE) is a clinical syndrome of neurological disorder caused by impaired cerebral oxygen delivery and blood flow around the time of birth. It affects approximately 1--5 per 1,000 term-born neonates \citep{long2007induced}. Therapeutic hypothermia reduces the risk of death and disability in infants with moderate-to-severe HIE \citep{silveira2015hypothermia}. However, mortality and adverse long-term outcomes, including neurodevelopmental delay and cognitive or motor impairment, still affect approximately $30\%$--$50\%$ of affected infants \citep{bao2025bonbid}. Accurate identification and localization of HIE-related brain injury are therefore important for quantitatively characterizing lesion severity and for understanding disease progression \citep{vermeulen2008diffusion, martinez2011predicting, chang2020predictive, bobba2023brain}.

Diffusion-weighted imaging (DWI) has been applied for detecting acute HIE-related injury, because abnormalities may appear earlier and more observable than signal changes on conventional T1- or T2-weighted MRI \citep{triulzi2006patterns,bano2017neonatal}. Cytotoxic edema restricts water diffusion, typically producing increased signal intensity on DWI and decreased apparent diffusion coefficient (ADC) values \citep{onda2024quantification}. ADC maps thus provide quantitative information for identifying HIE-related injury and assessing its severity and prognosis \citep{bobba2023brain,bao2025bonbid}. 

Automated localization remains challenging because HIE lesions are often small, multifocal, diffuse, and heterogeneous in appearance. In the BONBID-HIE cohort, more than half of affected neonates had lesions occupying less than $1\%$ of the brain volume \citep{bao2025bonbid}, resulting in severe foreground--background imbalance and poorly defined lesion boundaries. Scanner field strength, acquisition protocol, developmental anatomy, and postnatal imaging time further complicate the distinction between pathology and normal neonatal variability. These characteristics also make manual delineation difficult. Quantitative lesion analysis typically requires experienced pediatric neuroradiologists, often using multi-reader consensus \citep{bao2025bonbid}, but lesion boundaries may remain ambiguous and dependent on reader judgment. Consequently, expert voxel-level annotation is labor intensive and difficult to scale.

Supervised deep-learning approaches have achieved encouraging performance for neonatal HIE lesion segmentation \citep{bao2025bonbid,rao2025spatial,almansour2025enhancing}. However, the scarcity of publicly available neonatal HIE data \citep{bao2025bonbid}, along with the substantial expertise and effort required for annotation, restricts the scalability and external generalizability of these methods. Unsupervised anomaly detection provides an alternative by learning a normative representation from healthy images and identifying deviations in a potentially pathological subject without directly optimizing on lesion labels \citep{schlegl2019f,fernando2021deep}. A common strategy is to reconstruct a pseudo-healthy counterpart of the observed image and interpret the reconstruction residual as abnormality \citep{schlegl2019f,wolleb2022diffusion,wyatt2022anoddpm}.

However, reconstruction-based anomaly detection must balance pathology removal against preservation of healthy subject-specific anatomy. Excessive corruption or insufficient conditioning may alter normal structures and generate false-positive residuals, whereas strong subject conditioning may reproduce pathological signals and reduce lesion contrast. This trade-off is particularly problematic for HIE because diffusion abnormalities may be subtle, spatially diffuse, and poorly separated from normal neonatal variability \citep{forbes2000neonatal}. 
 
Diffusion models have recently emerged as expressive generative priors for unsupervised medical anomaly detection \citep{wolleb2022diffusion,wyatt2022anoddpm,FROTSCHER2026103895}. Their multi noise level denoising process provides an opportunity to identify deviations from learned normal anatomy other than the residual of a single final reconstruction. However, reliably distinguishing subtle pathological signals from subject-specific anatomy, acquisition-related variability, and stochastic denoising errors remains challenging. 

In this work, we propose OpenMAP-HIE, an anatomy-conditioned, pathology-informed score-based diffusion framework for unsupervised neonatal HIE lesion localization (Fig.~\ref{fig:architecture}). A diffusion model learns a healthy neonatal ADC prior, and a ControlNet branch incorporates the observed image as a spatial condition to capture patient-specific anatomy. We derive anomaly evidence from discrepancies between the healthy-prior and subject-conditioned score estimates, implemented through their noise-prediction differences (Fig.~\ref{fig:discrepancy_concept}). These discrepancies characterize how subject-specific conditioning modifies the local denoising response of the
learned healthy prior. Both branches evaluate the same noisy input at each noise level, with inputs generated directly from the observed image rather than propagated through a sequential reverse-diffusion chain. This formulation avoids dependence on a final pseudo-healthy reconstruction and the
propagation of errors between reverse-diffusion steps.

To improve pathological specificity, we incorporate a directional prior based on the characteristic ADC reduction associated with
acute HIE \citep{forbes2000neonatal,bobba2023brain,bao2025bonbid}. Specifically, we retain discrepancies whose corresponding
denoising estimates indicate a higher ADC under the healthy prior than under subject conditioning, and suppress discrepancies
in the opposite direction. The retained responses are aggregated across independently evaluated noise levels using a geometric mean to emphasize directionally consistent evidence. Population-derived ADC $Z$-maps
\citep{bao2025bonbid,rao2025spatial} are additionally investigated as an optional refinement, while the core framework requires only the subject's ADC image at inference.

The principal contributions of this work are:
\begin{itemize}
    \item We introduce an anatomy-conditioned diffusion framework that contrasts healthy-prior and subject-conditioned
    responses without using lesion masks as optimization targets.

    \item We formulate anomaly evidence through
    conditional--unconditional score discrepancies, implemented
    as noise-prediction differences and aggregated across independently evaluated noise levels, without requiring a sequential reverse-diffusion chain or a final pseudo-healthy reconstruction.
    
    \item We incorporate an HIE-specific directional prior that retains score discrepancies consistent with the expected ADC reduction of acute injury, adding pathology informed directionality to the
    score-based anomaly representation.
    
    \item We evaluate the framework on the public BONBID-HIE benchmark using voxel-ranking, lesion-overlap, and surface-distance metrics, and further assess population-derived ADC $Z$-map refinement as an optional extension.
\end{itemize}

The proposed ADC-only method achieved improved lesion overlap and surface localization relative to the evaluated unsupervised baselines, while the pathology-informed directional constraint provided substantial gains in lesion localization. Population-derived $Z$-map refinement further improved overlap and boundary accuracy when population-level normative information was available. The code of this work is available at \href{https://github.com/OishiLab/OpenMAP-HIE}{github.com/OishiLab/OpenMAP-HIE}.

\section{Related Works}\label{sec:related_works}
\subsection{Supervised Anomaly Detection}

Supervised medical image segmentation has advanced through encoder--decoder architectures such as U-Net \citep{ronneberger2015u} and UNet++
\citep{zhou2019unet++,rao2025spatial}. Recent HIE segmentation systems have extended to three-dimensional U-Net variants, SegResNet, Swin-UNETR \citep{hatamizadeh2021swin, almansour2025enhancing}, ensembles, and loss functions designed for highly imbalanced lesions \citep{bao2025bonbid}. These approaches learn lesion appearance directly from voxel-wise annotations. In contrast, our framework learns a healthy-image prior and derives lesion evidence without using lesion masks as optimization targets.

\subsection{Unsupervised Anomaly Detection}
Reconstruction-based anomaly detection learns normal image appearance and identifies abnormalities through discrepancies
between an input and its reconstructed counterpart. Autoencoders and variational autoencoders learn constrained representations of healthy images \citep{baur2021autoencoders}, while f-AnoGAN combines a GAN-based normal-image model with an encoder for efficient latent inference \citep{schlegl2019f}. Denoising autoencoders instead learn to recover healthy images
from artificially corrupted inputs, where appropriately designed coarse noise can improve anomaly detection while preserving
normal anatomical detail \citep{kascenas2022denoising}.
IterMask$^2$ combines iterative spatial mask refinement with high-frequency image information to improve reconstruction
of normal tissue and reduce false-positive responses \citep{liang2024itermask}.

These methods depend on reconstruction errors being larger in pathological regions than in healthy regions. However, normal anatomy may be reconstructed imperfectly, while sufficiently
expressive models may partially reproduce pathological structures. Consequently, reconstruction fidelity alone does not guarantee pathological specificity.

\subsection{Diffusion Model and Score Method}

Denoising diffusion probabilistic models (DDPMs) learn a data distribution by progressively perturbing images with noise and training a neural network to predict the noise of removal for the reverse process \citep{ho2020denoising}. Under the standard Gaussian perturbation formulation, noise prediction provides a parameterization of the score, the gradient of the perturbed data log density with respect to the noisy input
\citep{song2021scorebased}. This connection enables anomaly assessment through both reconstructed images and noise-level dependent noise predictions. Further, compared with GANs, diffusion models generally offer stable training and high-fidelity outputs \citep{dhariwal2021diffusion}, making them suitable for learning healthy anatomical priors.

Early diffusion-based methods primarily used pseudo-healthy reconstruction. AnoDDPM uses structured Simplex noise to obscure abnormalities at lower diffusion noise levels while retaining more image context \citep{wyatt2022anoddpm}. \citeauthor{wolleb2022diffusion} \citeyearpar{wolleb2022diffusion} combined deterministic Denoising Diffusion Implicit Model (DDIM) inversion \citep{song2020denoising} with classifier guidance \citep{dhariwal2021diffusion} to generate counterfactual healthy images. Patched diffusion models restrict corruption to local image regions and use surrounding anatomy as contextual information for denoising process, with pseudo-healthy reconstruction based on ensembles of predictions through sliding-window procedure \citep{behrendt2024patched}. 

Subsequent methods have attempted to restrict restoration to suspected pathological regions to avoid unnecessary modification of healthy anatomy. AutoDDPM first generates an anomaly-likelihood map and then uses masking, stitching, and joint-noise resampling to inpaint candidate abnormalities while harmonizing the reconstructed content with the original image \citep{bercea2023mask}. Temporal Harmonization for Optimal Restoration (THOR) instead constructs intermediate anomaly maps during reverse diffusion and uses them as implicit guidance to repeatedly restore presumed healthy regions from the original input \citep{bercea2024diffusion}. These methods are conceptually related to diffusion-based inpainting approaches such as RePaint \citep{lugmayr2022repaint}, in which known image regions are repeatedly reintroduced during sampling to constrain the generated result. 

Masking and local restoration have also been explored in latent diffusion frameworks, with introducing localized latent-space corruptions during
training as proxies for anomalous regions. MAD-AD applies forward diffusion only to randomly selected patches of healthy latent representations during training and jointly learns to reconstruct the clean latent representation and predict the region of corruption. At inference, its mask-prediction head identifies candidate anomalous patches, which are exclusively corrected during reverse diffusion while the remaining regions are preserved \citep{beizaee2025mad}. DeCo-Diff reformulates latent diffusion to learn a direction of deviation from normality. It progressively corrects anomalous latent patches while leaving normal patches unchanged and combines reconstruction discrepancies measured in the latent and image spaces to produce the final anomaly map \citep{beizaee2025correcting}.

Rather than producing a final pseudo-healthy reconstruction or iteratively refining a candidate inpainting mask, Aggregated Normative Diffusion (ANDi) derives anomaly evidence directly from deviations within the diffusion process \citep{FROTSCHER2026103895}. At each noise level, ANDi compares the analytical reverse-posterior mean conditioned on the observed image with the reverse mean
predicted by the diffusion model, both evaluated at the same noisy input. The resulting voxel-wise squared differences are aggregated across noise levels using a geometric mean. The voxel-wise squared deviations are aggregated across multiple sampled noise levels using a geometric mean. Consequently, ANDi does not require an additional inpainting stage or calculate its anomaly map only from a final reconstruction residual. Extending this reconstruction-free direction, InvAD instead uses DDIM inversion \citep{song2020denoising} to map image representations toward the noise
prior and evaluates deviations through Gaussian likelihood scoring in the resulting final latent space \citep{sakai2026invad}.

Our approach is most closely related to ANDi in its aggregation of discrepancies across noise levels. The key distinction is the comparison itself: we contrast subject-conditioned and learned healthy-prior score estimates, implemented through their noise-prediction differences, rather than comparing a model prediction with an analytical posterior target. We further filter the discrepancy signals using directional information from an HIE-specific
pathological prior.

\subsection{HIE Pathology Directional Prior}

Most reconstruction-based medical anomaly detection methods define abnormity from the difference between an observed image and a pseudo-healthy reconstruction, commonly using absolute or squared intensity residuals \citep{schlegl2019f,wolleb2022diffusion,wyatt2022anoddpm}. More recent diffusion-based approaches characterize abnormality through deviations from learned normal denoising dynamics \citep{bercea2024diffusion,FROTSCHER2026103895, beizaee2025mad} or gaussian prior likelihood \citep{sakai2026invad}. These disease-agnostic formulations are applicable to diverse anomaly types, but generally treat positive and negative intensity deviations symmetrically. Consequently, a large anomaly score does not have strong pathological interpretability and cannot properly distinguish the target pathology from the anatomical variance and imaging noise. 

This limitation is particularly relevant to neonatal HIE. During the acute diffusion-restriction phase, cytotoxic edema typically produces lower ADC values in injured tissue than in healthy tissue \citep{forbes2000neonatal,bobba2023brain,onda2024quantification,bao2025bonbid}. ADC thresholds and population-derived statistical maps have therefore been investigated for identifying and quantifying HIE-related injury \citep{bao2025bonbid}. However, the diffusion-based anomaly detection methods considered above do not explicitly incorporate the expected direction of the ADC abnormality into their multi noise-level discrepancy estimation. Our method incorporates this pathological information by retaining only positive differences between subject-conditioned and healthy-prior unconditional noise predictions, corresponding to higher healthy-reference posterior than subject-conditioned reverse-mean estimates. Unlike a conventional unsigned reconstruction residual, this formulation suppresses denoising discrepancies that are large in magnitude but inconsistent with the characteristic ADC reduction of acute HIE.

\section{Methods}\label{sec:method}
\subsection{Diffusion Model}
Let $\mathbf{x}_0\sim q_{\mathrm{h}}(\mathbf{x})$ denote a healthy neonatal ADC slice sampled from the healthy distribution. A denoising diffusion probabilistic model (DDPM) \citep{ho2020denoising} progressively perturbs $\mathbf{x}_0$ through the forward Markov process:
\begin{equation}
q(\mathbf{x}_t|\mathbf{x}_{t-1}) = \mathcal{N}
\left(\mathbf{x}_t;
\sqrt{1-\beta_t}\mathbf{x}_{t-1},
\beta_t\mathbf{I}
\right),
\label{eq:forward_stepwise}
\end{equation}

where $t=1,\dots, T$ and $\beta_t\in(0,1)$ specifies the variance schedule and
$\alpha_t=1-\beta_t$. Defining
$\bar{\alpha}_t=\prod_{s=1}^{t}\alpha_s$, a perturbed image at an arbitrary noise level $t$ can be sampled directly as
\begin{equation}
x_t = \sqrt{\bar{\alpha}_t}\mathbf{x}_0
+
\sqrt{1-\bar{\alpha}_t}\boldsymbol{\epsilon},
\qquad
\boldsymbol{\epsilon}\sim
\mathcal{N}(\mathbf{0},\mathbf{I}).
\label{eq:forward_markov}
\end{equation}
The reverse transition is parameterized as
\begin{equation}
p_\theta (x_{t-1}|x_t)=\mathcal{N}
\left(
\mathbf{x}_{t-1};
\boldsymbol{\mu}_\theta(\mathbf{x}_t,t),
\sigma_t^2\mathbf{I}
\right),
\label{eq}
\end{equation}
where $\theta$ denotes the parameters of the healthy-prior denoiser and $\sigma_t^2$ is determined by the adopted variance schedule. Under the noise-prediction parameterization, the reverse mean is
\begin{equation}
\boldsymbol{\mu}_\theta(\mathbf{x}_t, t) = \frac{1}{\sqrt{\alpha_t}} \left( \mathbf{x}_t - \frac{\beta_t}{\sqrt{1 - \bar{\alpha}_t}} \boldsymbol{\epsilon}_\theta(\mathbf{x}_t, t) \right),
\end{equation}

Following DDPM \citep{ho2020denoising} and using the Evidence Lower Bound (ELBO) loss, we train with minimizing the noise prediction loss:
\begin{equation}
    \mathcal{L}(\epsilon_\theta) := \mathbb{E}_{x_0\sim q(x_0), \epsilon\sim \mathcal{N}(0, \mathbf{I})}[\|\epsilon_\theta(x_t, t) - \epsilon\|^2]
\end{equation}
, where $x_t$ can be expressed with $x_0$ based on Eq. \eqref{eq:forward_markov}.

 Based on the Stochastic Differential Equation (SDE) formulation for the diffusion process \citep{song2021scorebased}, the noise prediction model $\epsilon_\theta$ is equivalent to the score function:
\begin{equation}
\epsilon_\theta(x_t, t)\propto -\nabla_{x_t} \log q(x_t|t)
\label{eq:score_def}
\end{equation}
, which points toward the higher density for the observed noised sample $x_t$ at time $t$.

\subsection{Spatial Conditioning through ControlNet}
To preserve the subject level anatomical features and reduce hallucination from the reverse process, we learn the conditional diffusion process with $\epsilon_{\theta,\phi}(x_t,t|c=x_0)$ using ControlNet \citep{zhang2023adding}. For training, we learn with:
\begin{equation}
    \mathcal{L}(\epsilon_{\theta, \phi}) := \mathbb{E}_{x_0\sim q(x_0), \epsilon\sim \mathcal{N}(0, \mathbf{I})}[\|\epsilon_{\theta,\phi}(x_t, t|c) - \epsilon\|^2]
\end{equation}

To prevent the model from ignoring its pre-trained generative priors, having the injected ControlNet signals to completely dominate the core diffusion branch and learning the shortcut of directly using condition $c=x_0$ to explicitly calculate $\epsilon_t$, we adopt the conditioning dropout strategy outlined in \cite{ho2021classifierfree} and \cite{zhang2023adding}. Specifically, during training, we randomly replace the conditioning inputs with a null condition $\varnothing$ into the model using a dropout probability of $p_{\text{drop}}=0.1$. Thus, we have an alternative condition pipeline during training:
\begin{equation}
     c= \Tilde{c} = \begin{cases}
        \varnothing \text{\quad , with probability }p_{\text{drop}}\\
        x_0  \text{\quad , otherwise}
    \end{cases}
\end{equation}
The base U-Net is pretrained on healthy images and remains frozen during ControlNet training, providing a fixed healthy-prior backbone. We use the null-conditioned prediction as the unconditional reference response:
\begin{equation}
\epsilon_{\theta,\phi}(x_t,t\mid c=\varnothing)
\simeq
\epsilon_{\theta}(x_t,t).
\label{eq}
\end{equation}
, which retains the pretrained healthy-prior backbone.

\subsection{Classifier-Free Guidance and Interpretation of the Conditional--Unconditional Discrepancy}
\label{sec:cfg_interpretation}

At inference, classifier-free guidance (CFG) \citep{ho2021classifierfree} provides a convenient formulation for combining the healthy-prior reference and subject-conditioned denoising responses. The guided noise prediction is defined as
\begin{equation}
\boldsymbol{\epsilon}^{\mathrm{CFG}}
(\mathbf{x}_t,t\mid\mathbf{c})
=
\boldsymbol{\epsilon}^{\mathrm{uncond}}
(\mathbf{x}_t,t)
+
w
\left[
\boldsymbol{\epsilon}^{\mathrm{cond}}
(\mathbf{x}_t,t\mid\mathbf{c})
-
\boldsymbol{\epsilon}^{\mathrm{uncond}}
(\mathbf{x}_t,t)
\right],
\label{eq:cfg}
\end{equation}
where $w\geq 0$ denotes the guidance scale, controlling the contribution of the subject-specific condition.

The unconditional prediction is obtained from the ControlNet-augmented model using a null condition,
\begin{equation}
\boldsymbol{\epsilon}^{\mathrm{uncond}}
(\mathbf{x}_t,t)
=
\boldsymbol{\epsilon}_{\theta,\phi}
(\mathbf{x}_t,t\mid\mathbf{c}=\varnothing).
\label{eq:cfg_uncond}
\end{equation} 

The conditional prediction is obtained by providing the observed ADC image as the spatial condition,
\begin{equation}
\boldsymbol{\epsilon}^{\mathrm{cond}}
(\mathbf{x}_t,t\mid\mathbf{c})
=
\boldsymbol{\epsilon}_{\theta,\phi}
(\mathbf{x}_t,t\mid\mathbf{c}=\mathbf{x}_0).
\label{eq:cfg_cond}
\end{equation}
We therefore use
$\boldsymbol{\epsilon}^{\mathrm{uncond}}$
as the healthy-reference response, whereas
$\boldsymbol{\epsilon}^{\mathrm{cond}}$
represents the denoising response after incorporating subject-specific information from the observed image.

The effect of the CFG scale can be understood directly from Eq.~\eqref{eq:cfg}:
\begin{itemize}
    \item When $w=0$, the guided prediction reduces to the null-conditioned healthy-reference response.
    \item For $0<w<1$,
\begin{equation}
\boldsymbol{\epsilon}^{\mathrm{CFG}}
=
(1-w)\boldsymbol{\epsilon}^{\mathrm{uncond}}
+
w\boldsymbol{\epsilon}^{\mathrm{cond}},
\label{eq:cfg_interp}
\end{equation}
which interpolates between the healthy-reference and subject-conditioned responses. Smaller values of $w$ reduce the influence of the observed subject and therefore allow stronger restoration toward the learned healthy distribution. However, this process may also reduce subject-specific anatomical fidelity and introduce reconstruction changes unrelated to pathology.
\item When $w=1$, the guided prediction becomes the standard conditional prediction. Importantly, the healthy prior is still retained through the frozen pretrained U-Net, while the ControlNet branch contributes its learned subject-specific residual. 
\item For $w>1$, CFG extrapolates beyond the null-conditioned healthy-reference response, along the conditional--unconditional direction. This formulation highlights the conditional--unconditional discrepancy
\begin{equation}
\boldsymbol{\delta}_t
=
\boldsymbol{\epsilon}^{\mathrm{cond}}
(\mathbf{x}_t,t\mid\mathbf{c}=\mathbf{x}_0)
-
\boldsymbol{\epsilon}^{\mathrm{uncond}}
(\mathbf{x}_t,t),
\label{eq:cfg_difference}
\end{equation}
which quantifies how the predicted denoising response changes when the observed subject image is introduced as spatial conditioning. Since the noise prediction is proportional to the negative score of the perturbed image distribution, as described in Eq.~\eqref{eq:score_def}, $\boldsymbol{\delta}_t$ can equivalently be interpreted as a local difference between the subject-conditioned and healthy-reference denoising fields.

In image regions whose appearance is compatible with the learned healthy distribution, the subject condition is expected to require only a limited modification of the healthy-reference denoising response, resulting in a relatively small discrepancy. In contrast, image characteristics that are poorly represented by the healthy prior may induce a larger conditional correction. Such deviations may arise from pathology, but may also reflect normal anatomical variation, registration differences, acquisition-related intensity variation, or imaging artifacts. The conditional--unconditional discrepancy therefore characterizes deviation from the learned healthy denoising behavior, as illustrated in Fig.~\ref{fig:discrepancy_concept}.

Repeated extrapolation with $w>1$ progressively bias the reconstruction toward features supported by the subject-specific condition. In a pathological input, this may result in stronger preservation or emphasis of lesion-associated appearance in the final reconstruction, which motivates the ``lesion-emphasizing'' CFG setting and conversely ``healthy-restoration'' for $w<1$ evaluated in the reconstruction-based ablation in Sec.~\ref{subsec:ablation_component}.

\end{itemize}  

However, at a fixed noisy input $\mathbf{x}_t$, CFG does not introduce additional discriminative information into this discrepancy, since
\begin{equation}
\boldsymbol{\epsilon}^{\mathrm{CFG}}
-
\boldsymbol{\epsilon}^{\mathrm{uncond}}
=
w\boldsymbol{\delta}_t.
\label{eq:cfg_scale_difference}
\end{equation}
Thus, changing $w$ only scales the conditional--unconditional direction at a given noise level. Although the guidance scale can substantially alter a complete reverse-diffusion trajectory and therefore affect reconstruction-based anomaly maps, the underlying discrepancy
$\boldsymbol{\delta}_t$
itself is independent of the selected CFG scale. In the following section, we therefore use the conditional--unconditional discrepancy directly and introduce HIE pathology informed directional information to determine which components of the deviation are biologically consistent with acute diffusion restriction.

\subsection{Pathology-Informed Directional Discrepancy and Multi Noise-Level Accumulation}
\label{subsec:discrepancy_score_accum}

Acute HIE injury is typically associated with restricted water diffusion and reduced ADC values relative to healthy brain tissue \citep{onda2024quantification}. We incorporate this characteristic as a directional pathological prior when comparing the subject-conditioned denoising response with the healthy-reference response.

For a given noisy input $\mathbf{x}_t$, we consider two reverse-diffusion responses: a healthy-prior branch and a subject-conditioned branch. The healthy-prior branch corresponds to the null-conditioned prediction,
\begin{equation}
\boldsymbol{\epsilon}^{\mathrm{prior}}
(\mathbf{x}_t,t)
\equiv
\boldsymbol{\epsilon}^{\mathrm{uncond}}
(\mathbf{x}_t,t),
\label{eq:prior_uncond_equiv}
\end{equation}
as defined in Eq.~\eqref{eq:cfg_uncond}, whereas the subject-conditioned branch uses
$\boldsymbol{\epsilon}^{\mathrm{cond}}(\mathbf{x}_t,t\mid\mathbf{c}=\mathbf{x}_0)$
from Eq.~\eqref{eq:cfg_cond}.

For each branch $b\in\{\mathrm{prior},\mathrm{cond}\}$, the corresponding reverse transition can be written as
\begin{equation}
\hat{\mathbf{x}}_{t-1}^{b}
=
\boldsymbol{\mu}^{b}_t
+
\sigma_t\mathbf{z}^{b}_t,
\qquad
\mathbf{z}^{b}_t\sim\mathcal{N}(\mathbf{0},\mathbf{I}),
\label{eq:reverse_sample_branches}
\end{equation}
where
\begin{equation}
\boldsymbol{\mu}^{b}_t
=
\frac{1}{\sqrt{\alpha_t}}
\left(
\mathbf{x}_t
-
\frac{\beta_t}{\sqrt{1-\bar{\alpha}_t}}
\boldsymbol{\epsilon}^{b}(\mathbf{x}_t,t)
\right).
\label{eq:reverse_mean_branches}
\end{equation}

To isolate the effect of subject-specific conditioning, the two branches are evaluated from the same noisy input $\mathbf{x}_t$ using the same stochastic realization $\mathbf{z}_t$.

The stochastic terms therefore cancel when the two reverse estimates are compared:
\begin{equation}
\hat{\mathbf{x}}_{t-1}^{\mathrm{prior}}
-
\hat{\mathbf{x}}_{t-1}^{\mathrm{cond}}
=
\boldsymbol{\mu}^{\mathrm{prior}}_t
-
\boldsymbol{\mu}^{\mathrm{cond}}_t.
\label{eq:shared_noise_difference}
\end{equation}

Substituting Eq.~\eqref{eq:reverse_mean_branches}, the reverse-mean difference can be expressed directly in terms of the corresponding noise predictions:
\begin{equation}
\boldsymbol{\mu}^{\mathrm{prior}}_t
-
\boldsymbol{\mu}^{\mathrm{cond}}_t
=
\frac{\beta_t}
{\sqrt{\alpha_t(1-\bar{\alpha}_t)}}
\left(
\boldsymbol{\epsilon}^{\mathrm{cond}}_t
-
\boldsymbol{\epsilon}^{\mathrm{prior}}_t
\right).
\label{eq:reverse_mean_difference}
\end{equation}
Since
$\boldsymbol{\epsilon}^{\mathrm{prior}}
\equiv
\boldsymbol{\epsilon}^{\mathrm{uncond}}$,
this can equivalently be written as
\begin{equation}
\begin{aligned}
&\boldsymbol{\mu}^{\mathrm{prior}}_t
-
\boldsymbol{\mu}^{\mathrm{cond}}_t
=
\kappa_t
\left(
\boldsymbol{\epsilon}^{\mathrm{cond}}_t
-
\boldsymbol{\epsilon}^{\mathrm{uncond}}_t
\right),\\
&\text{where }
\kappa_t
=
\frac{\beta_t}
{\sqrt{\alpha_t(1-\bar{\alpha}_t)}}.
\end{aligned}
\label{eq:reverse_mean_epsilon_relation}
\end{equation}
For all valid diffusion timesteps, $\kappa_t>0$. 

In a healthy region, both branches are expected to generate similar reverse estimates because the observed appearance is compatible with the learned healthy distribution. In an HIE lesion, however, the subject-conditioned branch is expected to preserve more of the abnormally low ADC signal present in the observed image, while the unconditional branch favors restoration toward a healthy-compatible appearance with a relatively higher ADC value. Therefore, for a voxel \(i\) within a low-ADC lesion, we expect
\begin{equation}
\mu^{\mathrm{prior}}_{t,i}
-
\mu^{\mathrm{cond}}_{t,i}
>0.
\end{equation}

Accordingly, we define the directional discrepancy score at reverse step \(t\) as
\begin{equation}
d_{t,i}
=
\left[
\mu^{\mathrm{prior}}_{t,i}
-
\mu^{\mathrm{cond}}_{t,i}
\right]_{+},
\quad
[u]_{+}=\max(u,0),
\label{eq:directional_discrepancy}
\end{equation}
where the positive-part operator retains only discrepancies consistent with restoration from a lower subject-conditioned ADC value toward a higher healthy-prior ADC value. Using Eq.~\eqref{eq:reverse_mean_epsilon_relation}, this discrepancy can be further expressed in terms of the predicted noise:
\begin{equation}
\begin{aligned}
d_{t,i}
&=
\left[
\kappa_t
\left(
\epsilon^{\mathrm{cond}}_{t,i}
-
\epsilon^{\mathrm{uncond}}_{t,i}
\right)
\right]_{+}\\
&\propto\left[
\epsilon^{\mathrm{cond}}_{t,i}
-
\epsilon^{\mathrm{uncond}}_{t,i}
\right]_{+}
\label{eq:directional_noise_discrepancy}
\end{aligned}
\end{equation}
, since $\kappa_t > 0$.

This formulation is directly related to the CFG representation introduced in Sec.~\ref{sec:cfg_interpretation}. From Eq.~\eqref{eq:cfg}, for $w>0$, CFG therefore preserves the sign of the underlying conditional--unconditional discrepancy while scaling its magnitude. Because our anomaly score is defined directly from
$\boldsymbol{\epsilon}^{\mathrm{cond}}
-
\boldsymbol{\epsilon}^{\mathrm{uncond}}$,
it does not require selecting a CFG scale. Similarly, although the reverse-mean discrepancy in Eq.~\eqref{eq:reverse_mean_epsilon_relation} contains the timestep-dependent coefficient $\kappa_t$, this coefficient does not alter the sign used by the pathological directional constraint. We therefore use the noise-prediction discrepancy itself as the operational anomaly representation rather than explicitly weighting it by $\kappa_t$.

A subtle lesion may produce only a weak discrepancy at any individual noise level. We therefore aggregate the directional evidence across a selected subset of independently evaluated noise levels
$\mathcal{T}\subseteq\{1,\ldots,T\}$.
For each $t\in\mathcal{T}$, a noisy input is constructed directly from the same observed image $\mathbf{x}_0$ according to Eq.~\eqref{eq:forward_markov}, with no state propagated from one evaluated noise level to another. The accumulated anomaly score at voxel $i$ is defined using the geometric mean \citep{FROTSCHER2026103895}:
\begin{equation}
A_i
=
\exp
\left(
\frac{1}{|\mathcal{T}|}
\sum_{t\in\mathcal{T}}
\log
\left(
d_{t,i} + \zeta_{\mathrm{stab}}
\right)
\right),
\label{eq:accumulated_discrepancy}
\end{equation}
where $\zeta_{\mathrm{stab}}>0$ is a small numerical stability constant introduced to avoid taking the logarithm of zero.

The positive-part operation is applied before accumulation so that discrepancies opposite to the expected low-ADC lesion direction do not cancel directionally consistent responses. Compared with an arithmetic mean, geometric aggregation reduces the influence of isolated large discrepancies and favors voxels that exhibit consistently elevated directional responses across multiple corruption levels \citep{FROTSCHER2026103895}. Because each noise level is evaluated independently from the original subject image, the accumulated score reflects persistent disagreement between the subject-conditioned and healthy-reference denoising fields without requiring a sequential reverse-diffusion trajectory or a final pseudo-healthy reconstruction.

The directional constraint improves pathological relevance but does not by itself establish lesion specificity. Normal anatomical variation, partial-volume effects, registration differences, acquisition-related intensity variation, and imaging artifacts may also produce positive discrepancies. The resulting score is therefore interpreted as evidence of a persistent deviation from the learned healthy denoising behavior that is directionally consistent with acute HIE-related ADC reduction, rather than as a calibrated lesion probability.

\subsection{Spatial Smoothing of the Anomaly Map}

Although accumulation across multiple noise levels enhances persistent
conditional--unconditional discrepancies, the resulting anomaly map may
contain spatially isolated high-frequency responses. As following \cite{FROTSCHER2026103895}, we apply a
mean filtering to improve the local spatial coherence of the accumulated score.

Let \(A_i\) denote the accumulated directional discrepancy at voxel \(i\).
The spatially regularized anomaly score is defined as
\begin{equation}
\widetilde{A}_i
=
\frac{1}{|\mathcal{N}_k(i)|}
\sum_{j\in\mathcal{N}_k(i)} A_j,
\label{eq:mean_filter}
\end{equation}
where \(\mathcal{N}_k(i)\) denotes the local \(k\times k\) neighborhood centered at voxel
\(i\), and \(k\) is the mean-filter kernel size. The same filter is applied
to all subjects before thresholding the continuous anomaly map.

Mean filtering reduces isolated high-valued responses by averaging them with low-valued neighboring voxels, whereas spatially coherent regions remain relatively elevated because their neighboring voxels exhibit consistently high discrepancy scores. However, an excessively
large kernel may dilute focal lesion responses and spread elevated anomaly
scores into adjacent healthy tissue, with increasing false positive
detections. We use \(k=5\) in the final framework,
as selected using the validation set. The sensitivity of the method to the
kernel size is evaluated in Table~\ref{tab:mean_filter_ablation}.

\subsection{Population Z-Map Guided Post-processing Lesion Refinement}
\label{subsec:method_postprocessing}
The diffusion-derived anomaly map characterizes subject-specific disagreement between the conditional and unconditional noise predictions. Although this discrepancy provides complementary evidence of abnormal tissue, it may also respond to normal anatomical variation, tissue boundaries, registration errors, and acquisition-related artifacts. We therefore combine the diffusion anomaly score with a population-normalized ADC $Z$-map \citep{bao2025bonbid}, which provides an independent spatial measure of deviation from the healthy reference population. Because acute HIE lesions are generally associated with restricted diffusion and reduced ADC values \citep{onda2024quantification}, strongly negative $Z$-scores provide biologically consistent evidence for lesion localization \citep{bao2025bonbid, rao2025spatial}.

Let \(A_i\) denote the accumulated directional noise-prediction discrepancy at voxel \(i\), as defined in Eq.~\eqref{eq:accumulated_discrepancy}, and let \(Z_i\) denote the corresponding population ADC \(Z\)-score. All operations are performed independently for each subject:
\begin{equation}
\Omega
=
\left\{
i:
A_i\ \text{and}\ Z_i\ \text{are finite}
\right\}.
\end{equation}

\paragraph{High-confidence low-ADC core.}
A high-confidence lesion core is first identified solely from the population $Z$-map:
\begin{equation}
C_{\mathrm{raw}}
=
\left\{
i\in\Omega:
Z_i\leq -2.0
\right\},
\label{eq:raw_zmap_core}
\end{equation}
as following \cite{rao2025spatial} where $-2.0$ is an consistent threshold for lesion on $Z$-map.

Three-dimensional connected-component analysis was performed over the entire volume binary mask. Voxels sharing a face, edge, or corner were considered connected (26-connectivity), and connected regions containing fewer than 150 voxels were removed:
\begin{equation}
C
=
\bigcup_{k:|C_k|\geq150}C_k,
\qquad
{C_k}
=
\operatorname{CC}_{26}(C_{\mathrm{raw}}),
\label{eq:clean_zmap_core}
\end{equation}
where \(\operatorname{CC}_{26}(\cdot)\) denotes 26-connected component decomposition. The size constraint of each lesion component suppresses isolated negative $Z$-score responses that are unlikely to represent spatially coherent HIE lesions.

\paragraph{Diffusion-supported component recovery.}
The diffusion anomaly map is subsequently used to recover complementary lesion components that may not be included in the high-confidence \(Z\)-map core. A subject-specific high-anomaly threshold is defined as $\tau_A
=
Q_{99}\left(A;\Omega\right),
$
corresponding to the 99th percentile of the normalized anomaly scores. Candidate voxels must simultaneously exhibit a high diffusion anomaly score and a negative population deviation:
\begin{equation}
R
=
\left\{
i\in\Omega:
A_i\geq\tau_A
\ \land
Z_i\leq-1.95
\right\}.
\label{eq:anomaly_rescue_candidates}
\end{equation}
Thus, the diffusion score alone cannot introduce a candidate voxel without supporting evidence of reduced ADC relative to the healthy population.

The candidate mask \(R\) is decomposed into 26-connected components,
\begin{equation}
\{R_j\}_{j=1}^{J}
=
\operatorname{CC}_{26}(R).
\end{equation}
A candidate component \(R_j\) is retained only if its size is larger than 10 voxels, i.e. $10\leq|R_j|$ and it is spatially adjacent to the cleaned \(Z\)-map core:
\begin{equation}
R_j
\cap
\operatorname{Dilate}_{26}(C,1)
\neq
\varnothing,
\label{eq:rescue_spatial_constraint}
\end{equation}
where \(\operatorname{Dilate}_{26}(C,1)\) denotes a one-voxel three-dimensional dilation of \(C\) using 26-connectivity. When a candidate component satisfies both constraints, the entire connected component is included in the prediction. This component-level operation allows the method to recover spatially coherent lesion extensions while preventing distant high-anomaly regions from being detected independently. The set of accepted components is therefore
\begin{equation}
\mathcal{J}_{\mathrm{acc}}
=
\left\{
j:
10\leq|R_j|
\ \land
R_j\cap\operatorname{Dilate}_{26}(C,1)\neq\varnothing
\right\}.
\label{eq:accepted_rescue_components}
\end{equation}
The final binary lesion map for this current subject is defined as
\begin{equation}
M
=
C
\cup
\bigcup_{j\in\mathcal{J}_{\mathrm{acc}}}R_j.
\label{eq:final_zmap_refined_prediction}
\end{equation}

This conservative fusion combines two complementary representations of abnormality. The population \(Z\)-map provides the primary localization of spatially coherent low-ADC tissue, whereas the diffusion discrepancy supplies subject-specific evidence for recovering nearby lesion components. No manual lesion annotation is required to generate the final prediction at inference.

\subsection{Dataset and Preprocessing}

We used the BONBID-HIE dataset \citep{bao2025bonbid} for model development and evaluation. The dataset was divided into 85 training subjects, 4 validation subjects, and 44 test subjects, as following its predefined partitions, without performing an additional subject-level split and with no subject overlap among the three partitions. All 133 subjects in the selected release were included, with no subject-level exclusions. Each subject included an three-dimensional ADC map and its corresponding population-normalized $Z$-map. To account for variations in voxel spacing and spatial misalignment across scans, all volumes were affine registered to the JHU-neonate-SS atlas template \citep{oishi2011multi} using ANTsPy \citep{tustison2021antsx}. At the end, all scans were resampled to a common matrix size \(256\times256\times128\) and voxel size $0.58\times 0.58\times 0.74$ $mm^3$, with nearest neighbor interpolation, as following \cite{rao2025spatial}. The intensity of each volume went through its independent 99th percentile normalization.

The diffusion model was trained using two-dimensional axial ADC slices. To construct a training set representing non-lesion brain tissue, we examined the corresponding lesion mask for each slice from the training subjects and retained only slices containing no annotated lesion voxels. The lesion masks were used exclusively for training-slice selection and were not provided to the model as inputs or optimization targets. The $Z$-maps were not used to train the diffusion model and were introduced only during the post-processing stage described in Section~\ref{subsec:method_postprocessing}. We have 6042 slices of non-lesion axial brain slices for training and 4838 slices of lesion axial brain slices from the training subjects for the hyperparameter selection for the post-processing stage.

The predefined validation partition was used to monitor loss and model performance during training to reduce overfitting and to select the mean-filter kernel size. Evaluation was performed over complete 3D volumes from the 44 test subjects.

\subsection{Network Architecture and Implementation Details}

The two-stage training procedure is summarized in Algorithm~\ref{alg:two_stage_training}, and the discrepancy-based inference procedure is described in Algorithm~\ref{alg:discrepancy_inference}. The diffusion denoiser was implemented using a U-Net architecture \citep{ronneberger2015u}. The ControlNet conditioning branch was constructed using the encoder portion of the U-Net architecture, allowing the observed subject image to be encoded at multiple spatial resolutions \citep{zhang2023adding} (Fig.~\ref{fig:architecture}). As following \cite{zhang2023adding}, at each resolution level, the features produced by the ControlNet encoder were projected through a zero-initialized convolution and injected into the corresponding feature stream of the primary U-Net through a residual connection. Specifically, the fused feature representation at resolution level \(\ell\) was defined as
\begin{equation}
\widetilde{\mathbf{h}}_{\ell}
=
\mathbf{h}_{\ell}^{\mathrm{UNet}}
+
\mathcal{Z}_{\ell}
\left(
\mathbf{h}_{\ell}^{\mathrm{Control}}
\right),
\label{eq:controlnet_residual}
\end{equation}
where \(\mathbf{h}_{\ell}^{\mathrm{UNet}}\) and \(\mathbf{h}_{\ell}^{\mathrm{Control}}\) denote the primary U-Net and ControlNet feature maps, respectively, and \(\mathcal{Z}_{\ell}\) denotes the zero-initialized convolutional projection. Both the weights and biases of \(\mathcal{Z}_{\ell}\) were initialized to zero. Consequently, the ControlNet residual was initially zero and the pretrained U-Net was frozen throughout the second stage of training, allowing the pretrained U-Net to preserve its learned healthy denoising behavior at the beginning of the second training stage. As training progressed, the residual projections learned to introduce subject-specific spatial information and to adjust the pre-learned healthy prior.

Both training stages were optimized using AdamW \citep{loshchilov2017decoupled} with a batch size of 8. We employed a warm-up cosine-decay learning-rate schedule. During the first $5\%$ of the total optimization steps, the learning rate was increased from $2\times10^{-5}$ to the target learning rate of $1\times10^{-4}$. It was subsequently decayed according to a cosine schedule over the remaining optimization steps. An exponential moving average of the model parameters was maintained with a decay rate of $0.995$.

In the first stage, the U-Net diffusion model was trained for $K_1=300$ epochs to learn the healthy neonatal image distribution. In the second stage, the pretrained U-Net served as the healthy generative prior, and the ControlNet branch was trained for an additional $K_2=200$ epochs to incorporate subject-specific spatial conditioning. To enable both null-conditioned and subject-conditioned noise predictions, the conditioning image was randomly dropped with probability \(p_{\mathrm{drop}}=0.1\) during ControlNet training.

The diffusion process used \(T=1000\) noise steps with a linear variance schedule, in which \(\beta_t\) increased from \(\beta_1=1\times10^{-4}\) to \(\beta_T=0.02\). During training, the input images were normalized, and random horizontal flipping was applied with a probability of \(0.5\).

During inference, following the intermediate-noise strategy described in \citep{FROTSCHER2026103895}, we accumulated the conditional-unconditional discrepancy over diffusion timesteps $\mathcal{T} = \{75,76,\ldots,199\}.$ This interval was selected to balance corruption of the pathological signal with preservation of subject-specific anatomy. At very low noise levels, the lesion appearance remains largely preserved in \(\mathbf{x}_t\), providing insufficient perturbation for the healthy prior to reveal a clear discrepancy. Conversely, at substantially higher noise levels, excessive corruption removes subject-specific anatomical information and may introduce reconstruction differences unrelated to pathology. The selected intermediate timesteps weaken the lesion-related signal while retaining sufficient anatomical information for spatially meaningful comparison between the conditional and unconditional noise predictions.

\subsection{Evaluation Metrics}
\label{sec:evaluation_metrics}

We evaluated the predicted anomaly maps using complementary
voxel-ranking, volumetric-overlap, and surface-distance
metrics. Let $A_{s,i}$ denote the continuous anomaly score
at voxel $i$ of subject $s$, and let $G_s$ denote the
corresponding ground truth lesion mask. Given a threshold
$\lambda$, the binary prediction is $P_s(\lambda)=\{i:A_{s,i}>\lambda\}$. Each metric was first computed over the complete 3D volume of each subject, rather than independently on individual 2D slices, and was reported as the mean $\pm$ standard deviation across the test subjects.

\paragraph{Dice similarity coefficient and threshold selection.}
The Dice similarity coefficient measures volumetric
overlap between the predicted mask $P$ and the ground truth
mask $G$:
\begin{equation}
\operatorname{Dice}(P,G)
=
\frac{2|P\cap G|}{|P|+|G|},
\label{eq:dice}
\end{equation}
where $|\cdot|$ denotes the number of foreground voxels.
Dice ranges from 0 to 1, with higher values indicating
greater overlap.

For each evaluated method, we determined the \textit{ceiling Dice} by selecting a single threshold $\lambda^{*}$ that maximized the mean subject-wise Dice score on the test set:
\begin{equation}
\lambda^{*}
=
\operatorname*{arg\,max}_{\lambda\in\Lambda}
\frac{1}{N_{\mathrm{test}}}
\sum_{s=1}^{N_{\mathrm{test}}}
\operatorname{Dice}\bigl(P_s(\lambda),G_s\bigr),
\label{eq:ceiling_dice_threshold}
\end{equation}
where $\Lambda$ denotes the candidate threshold set and $N_{\mathrm{test}}$ is the number of test subjects. We reported the resulting ceiling Dice as the Dice score throughout the manuscript. The same threshold $\lambda^{*}$ was then applied to all test subjects to compute the following described surface-distance metrics without further optimization. These binary-segmentation metrics therefore describe performance at a Dice-optimal threshold selected using the test-set annotations.

\paragraph{Average precision.}
Average precision (AP) evaluates voxel-level ranking
by summarizing the precision--recall curve obtained
from the continuous anomaly scores:
\begin{equation}
\operatorname{AP}
=
\sum_k
\left(
\operatorname{Rec}_k-\operatorname{Rec}_{k-1}
\right)
\operatorname{Prec}_k,
\label{eq:average_precision}
\end{equation}
where $\operatorname{Prec}_k$ and $\operatorname{Rec}_k$
denote precision and recall at the $k$-th sorted threshold.
AP does not depend on the selected threshold
$\lambda^{*}$ and is particularly informative because
lesion voxels constitute a small proportion of the
evaluated brain volume. Higher AP indicates better ranking
of lesion voxels above non-lesion voxels.

\paragraph{Surface-distance metrics.}
For clarity, we omit the subject index below and
write $P=P_s(\lambda^{*})$ and $G=G_s$.
Let $S_P=\partial P$ and $S_G=\partial G$ denote
the predicted and ground truth lesion surfaces.
The distance from a surface voxel $x$ to a surface
$S$ is
\begin{equation}
d(x,S)=\min_{y\in S}\|x-y\|_2,
\label{eq:point_surface_distance}
\end{equation}
where coordinates are expressed in physical space,
accounting for voxel spacing. All distances are
reported in millimeters.

The \textit{symmetric Hausdorff distance} (HD) measures
the maximum surface disagreement:
\begin{equation}
\operatorname{HD}(P,G)
=
\max\left\{
\max_{x\in S_P}d(x,S_G),
\max_{y\in S_G}d(y,S_P)
\right\}.
\label{eq:hausdorff}
\end{equation}
Lower HD indicates a smaller maximum boundary error.

The \textit{mean average surface distance} (MASD) measures the mean bidirectional distance between the two lesion surfaces:
\begin{equation}
\begin{aligned}
\operatorname{MASD}(P,G)
&=
\frac{1}{2}
\left[
\bar d(S_P,S_G)+\bar d(S_G,S_P)
\right],\\
\bar d(S_1,S_2)
&=
\frac{1}{|S_1|}
\sum_{x\in S_1}d(x,S_2).
\end{aligned}
\label{eq:masd}
\end{equation}
Lower MASD indicates closer average agreement
between the predicted and reference boundaries.

\textit{Normalized surface distance} (NSD) measures the proportion
of surface voxels within a predefined dilated
tolerance $\tau$ of the corresponding surface:
\begin{equation}
\operatorname{NSD}_{\tau}(P,G)
=
\frac{
\displaystyle
\sum_{x\in S_P}
\mathbf{1}\!\left[d(x,S_G)\leq\tau\right]
+
\displaystyle
\sum_{y\in S_G}
\mathbf{1}\!\left[d(y,S_P)\leq\tau\right]
}{
|S_P|+|S_G|
},
\label{eq:nsd}
\end{equation}
where $\mathbf{1}[\cdot]$ is the indicator function.
We used a tolerance of $\tau=2.0$~mm in the resampled
image space, following \citet{rao2025spatial}.
NSD ranges from 0 to 1, with higher values indicating
better surface agreement within the specified tolerance.

\renewcommand{\algorithmicrequire}{\textbf{Input:}}
\renewcommand{\algorithmicensure}{\textbf{Output:}}
\begin{algorithm}[t]
\footnotesize
\caption{Two-stage training of the healthy-prior DDPM and ControlNet}
\label{alg:two_stage_training}
\begin{algorithmic}[1]

\Require Healthy ADC dataset $\mathcal{D}_{\mathrm{h}}$;
diffusion schedule $\{\beta_t\}_{t=1}^{T}$;
training iterations $K_1$ and $K_2$;
condition-dropout probability $p_{\mathrm{drop}}$
\Ensure Healthy-prior parameters $\theta$ and ControlNet parameters $\phi$

\State Define $\alpha_t=1-\beta_t$ and
$\bar{\alpha}_t=\prod_{s=1}^{t}\alpha_s$

\Statex
\Statex \textbf{Stage 1: Healthy-prior DDPM}

\State Initialize the unconditional denoiser
$\boldsymbol{\epsilon}_{\theta}$

\For{$k=1,\ldots,K_1$}
    \State Sample $\mathbf{x}_0\sim\mathcal{D}_{\mathrm{h}}$,
    $t\sim\mathcal{U}\{1,\ldots,T\}$, and
    $\boldsymbol{\epsilon}\sim\mathcal{N}(\mathbf{0},\mathbf{I})$
    \State Construct
    $\mathbf{x}_t=
    \sqrt{\bar{\alpha}_t}\mathbf{x}_0+
    \sqrt{1-\bar{\alpha}_t}\boldsymbol{\epsilon}$
    \State Compute
    $\mathcal{L}_{\mathrm{prior}}
    =
    \|
    \boldsymbol{\epsilon}
    -
    \boldsymbol{\epsilon}_{\theta}(\mathbf{x}_t,t)
    \|_2^2$
    \State Update $\theta$ using
    $\nabla_{\theta}\mathcal{L}_{\mathrm{prior}}$
\EndFor

\Statex
\Statex \textbf{Stage 2: ControlNet}

\State Freeze $\theta$ and initialize the ControlNet parameters $\phi$

\For{$k=1,\ldots,K_2$}
    \State Sample $\mathbf{x}_0\sim\mathcal{D}_{\mathrm{h}}$,
    $t\sim\mathcal{U}\{1,\ldots,T\}$, and
    $\boldsymbol{\epsilon}\sim\mathcal{N}(\mathbf{0},\mathbf{I})$
    \State Construct
    $\mathbf{x}_t=
    \sqrt{\bar{\alpha}_t}\mathbf{x}_0+
    \sqrt{1-\bar{\alpha}_t}\boldsymbol{\epsilon}$
    \State Set
    $\widetilde{\mathbf{c}}=\varnothing$
    with probability $p_{\mathrm{drop}}$;
    otherwise set $\widetilde{\mathbf{c}}=\mathbf{x}_0$
    \State Compute
    $\mathcal{L}_{\mathrm{cond}}
    =
    \|
    \boldsymbol{\epsilon}
    -
    \boldsymbol{\epsilon}_{\theta,\phi}
    (\mathbf{x}_t,t\mid\widetilde{\mathbf{c}})
    \|_2^2$
    \State Update only $\phi$ using
    $\nabla_{\phi}\mathcal{L}_{\mathrm{cond}}$
\EndFor

\State \Return $\theta,\phi$

\end{algorithmic}
\end{algorithm}

\begin{algorithm}[t]
\footnotesize
\caption{Multi noise-level conditional-unconditional score discrepancy inference}
\label{alg:discrepancy_inference}
\begin{algorithmic}[1]

\Require Patient ADC slice $\mathbf{x}_0$;
selected noise levels $\mathcal{T}$;
trained parameters $\theta,\phi$; small numerical stability constant $\zeta_{\mathrm{stab}}$; brain mask $\mathbf{B}$
\Ensure Diffusion-derived anomaly map $\mathbf{A}_{\mathrm{diff}}$

\State Set patient condition $\mathbf{c}\gets\mathbf{x}_0$
\State Initialize $\mathbf{A}\gets\mathbf{0}$

\ForAll{$t\in\mathcal{T}$}
    \State Sample
    $\boldsymbol{\epsilon}_t\sim\mathcal{N}(\mathbf{0},\mathbf{I})$
    \State $\mathbf{x}_t\gets
    \sqrt{\bar{\alpha}_t}\mathbf{x}_0+
    \sqrt{1-\bar{\alpha}_t}\boldsymbol{\epsilon}_t$
    \Comment{Direct perturbation}

    \State $\widehat{\boldsymbol{\epsilon}}_t^{\mathrm{prior}}
    \gets\boldsymbol{\epsilon}_{\theta, \phi}(\mathbf{x}_t,t|\varnothing)$

    \State $\widehat{\boldsymbol{\epsilon}}_t^{\mathrm{cond}}
    \gets\boldsymbol{\epsilon}_{\theta,\phi}
    (\mathbf{x}_t,t\mid\mathbf{c})$

    \State $\mathbf{d}_t\gets
    \widehat{\boldsymbol{\epsilon}}_t^{\mathrm{cond}}
    -\widehat{\boldsymbol{\epsilon}}_t^{\mathrm{prior}}$

    \State $\mathbf{A}\gets\mathbf{A}+\log \left([\mathbf{d}_t]_{+}+ \zeta_{\mathrm{stab}}\right)$
    \Comment{Pathological directional prior}
\EndFor

\State $\mathbf{A}_{\mathrm{diff}}\gets
\exp\bigl(\mathbf{A}/|\mathcal{T}|\bigr)\odot\mathbf{B}$
\Comment{Geometric mean}
\State \Return $\mathbf{A}_{\mathrm{diff}}$
\Comment{No sequential reverse chain}

\end{algorithmic}
\end{algorithm}

\subsection{Statistical Analysis}
\label{subsec:statistical_analysis}

Statistical comparisons were performed using two-sided paired Wilcoxon signed-rank tests, with each test subject serving as the paired unit. Dice was the primary outcome for comparisons between the proposed method and competing baselines, as well as for selected ablation analyses. Statistical significance was assessed at $\alpha=0.05$. Mean paired Dice differences were reported alongside $p$-values to describe the magnitude of the observed effects.

For methods evaluated using ceiling Dice, subject-level scores were computed at the same method-specific threshold selected to maximize mean test-set Dice. These statistical comparisons therefore characterize performance under the oracle threshold evaluation protocol rather than at independently selected operating thresholds.

\subsection{Alternative Discrepancy Formulations for Ablation}
\label{subsec:discrepancy_formulation_ablation}
To assess the design choices underlying the proposed discrepancy score, we
consider two alternative formulations. These formulations are evaluated only as ablation variants and are not used in the final framework. The evaluation of the ablation variants is in Table \ref{tab:score_chain_ablation}.

\paragraph{Reverse-chain accumulation.}
Our final method evaluates each selected noise level independently by directly sampling $x_t$ using Eq. \eqref{eq:forward_markov}. As an alternative, we construct a dependent reverse trajectory in which each
state is generated from the preceding reverse step:
\begin{equation}
\mathbf{x}_{t-1}^{\mathrm{chain}}
=
\boldsymbol{\mu}_{\theta}
(\mathbf{x}_{t}^{\mathrm{chain}},t)
+
\sigma_t\mathbf{z}_t,
\quad
\mathbf{z}_t\sim\mathcal{N}(\mathbf{0},\mathbf{I}).
\label{eq:chained_discrepancy}
\end{equation}
The discrepancy at each timestep is then evaluated on
\(\mathbf{x}_{t}^{\mathrm{chain}}\). This variant process tests whether preserving a
continuous reverse trajectory is preferable to independently examining matched
corruption levels of the observed image.

\paragraph{Forward-noise reference.}
We additionally replace the unconditional model prediction with the Gaussian
noise used in the forward process:
\begin{equation}
\begin{aligned}
    \widetilde{d}_t
&=
\boldsymbol{\epsilon}^{\mathrm{cond}}
(\mathbf{x}_t,t\mid c=\mathbf{x}_0)
-
\boldsymbol{\epsilon}_t,\\
\text{where } x_t &= \sqrt{\bar{\alpha}_t}x_0+\sqrt{1-\bar{\alpha}_t}\epsilon_t
\end{aligned}
\label{eq:forward_noise_discrepancy}
\end{equation}
This variant formulation tests whether anomaly evidence can be obtained from conditional noise prediction error alone, without using the learned unconditional response
as a healthy reference.
\section{Results}\label{sec:result}
\subsection{Unsupervised Benchmark Results}
\begin{figure*}
    \centering
    \includegraphics[width=1.0\linewidth]{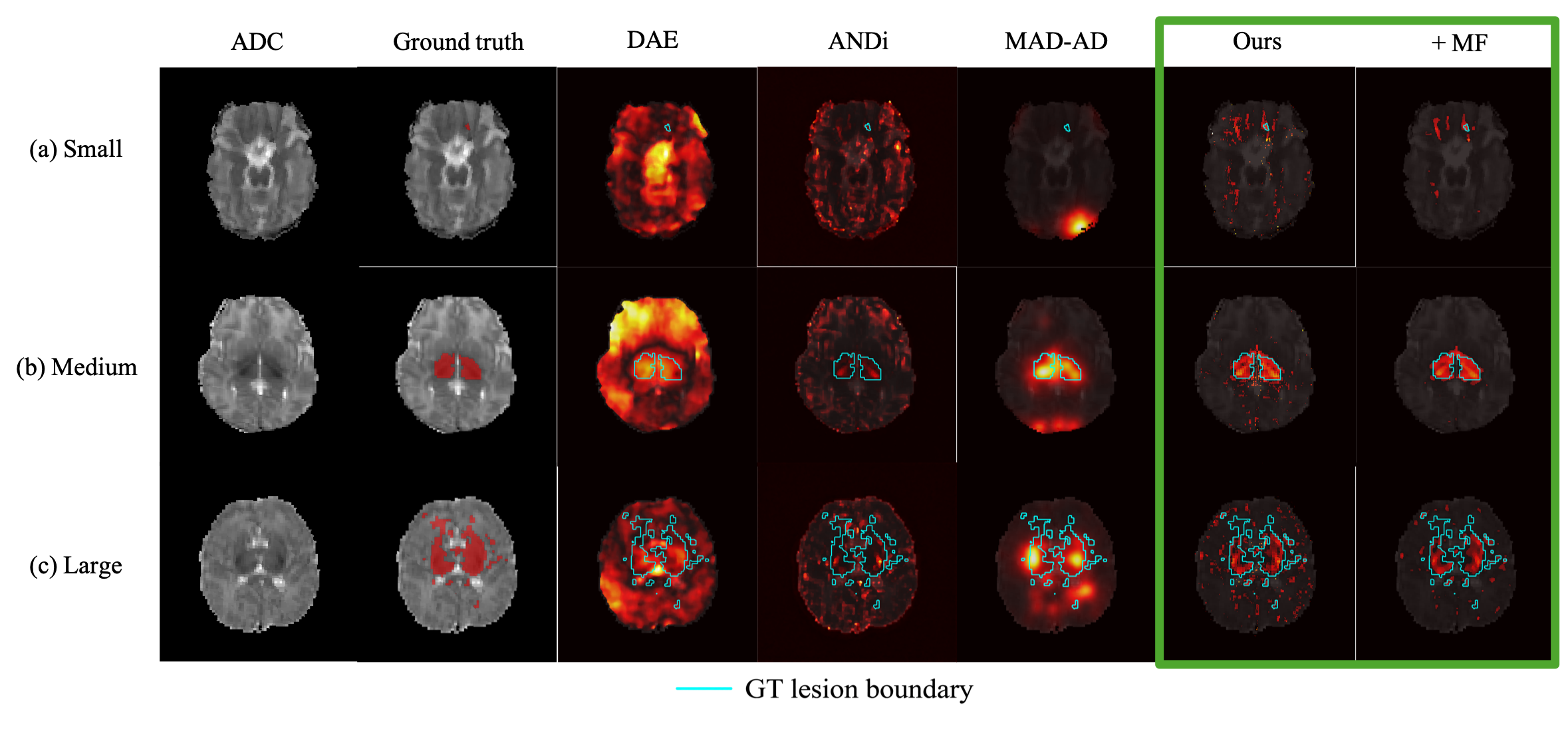}
    \caption{Qualitative comparison of unsupervised anomaly detection methods in
subjects with (a) small ($< 1\%$), (b) medium ($1\% - 5\%$), and (c) large ($\ge 5\%$) lesion burdens, defined by the subject-level lesion percentage.
Columns show the input ADC image, expert lesion annotation, and
anomaly maps from selected top-performing baselines, including DAE \citep{kascenas2022denoising}, ANDi \citep{FROTSCHER2026103895}, MAD-AD \citep{beizaee2025mad}, where the originally proposed smoothing methods from each paper have been applied. Our proposed method before
and after mean filtering (MF; $k=5$) is highlighted in green box.  
Red overlays in the ground truth (GT) column indicate annotated lesions,
and cyan contours delineate their boundaries on the anomaly maps.
Brighter colors indicate higher anomaly scores.}
    \label{fig:anomaly_demos}
\end{figure*}

\begin{figure}
    \centering
    \includegraphics[width=0.93\linewidth]{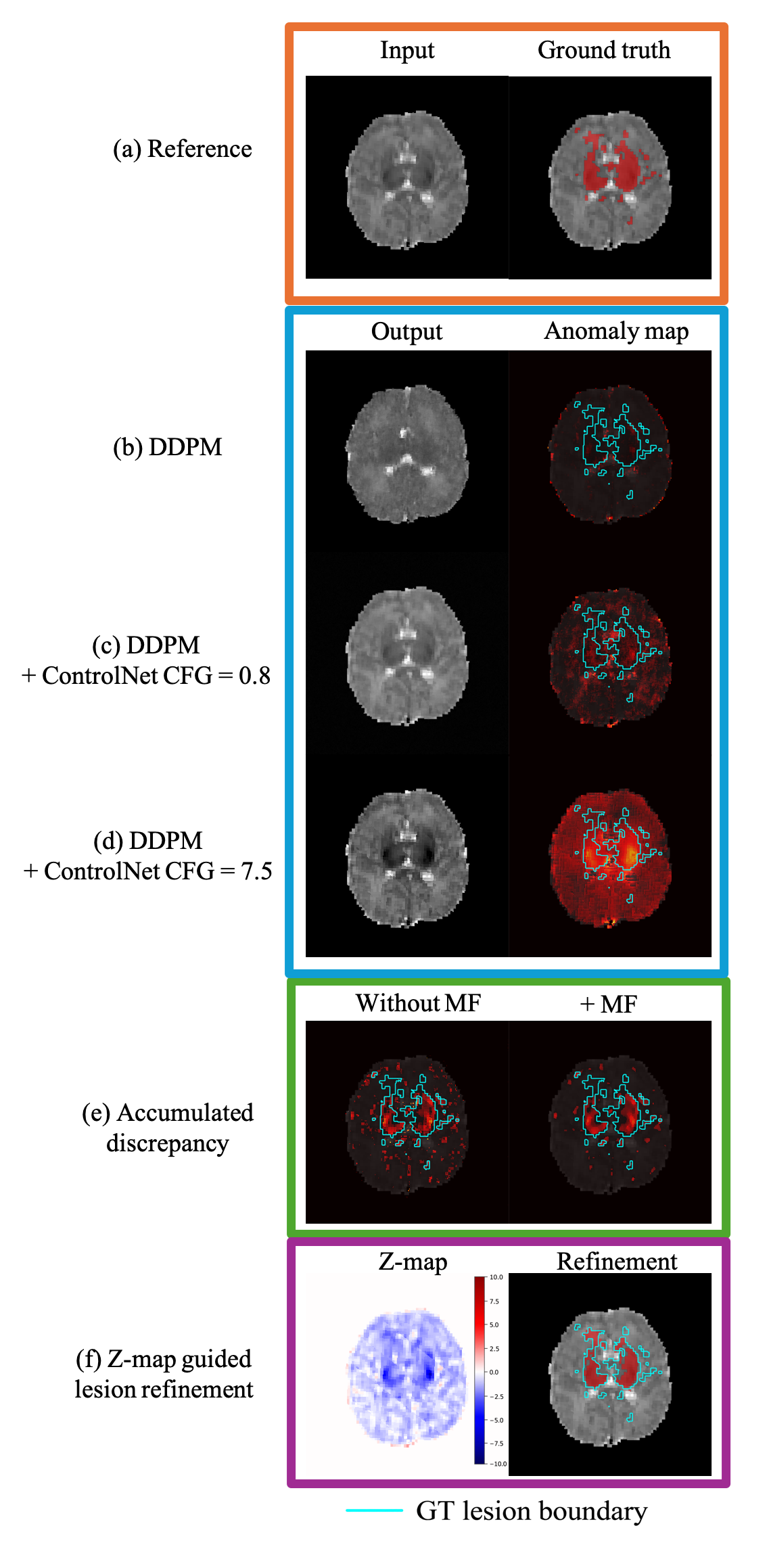}
    \caption{Qualitative comparison of reconstruction-based anomaly detection and
the proposed discrepancy-based approach. (a) Input ADC image and expert lesion annotation. (b--d) Reconstructed images and corresponding anomaly maps obtained using DDPM and DDPM with ControlNet at CFG scales of 0.8 and 7.5. (e) Anomaly maps produced by the proposed method before and after mean filtering (MF, $k=5$). (f) The corresponding population-derived Z-map for the input and the Z-map guided post-processing lesion refinement. Cyan contours delineate the expert-annotated lesion boundaries. All methods' results have been applied with HIE pathological informed sign direction.}
    \label{fig:ablation_demos}
\end{figure}

\begin{table*}[t]
\centering
\caption{Quantitative comparison of unsupervised anomaly detection methods for neonatal HIE lesion localization. Values are reported as mean $\pm$ standard deviation where available. The best and second-best results are shown in \textbf{bold} and \underline{underlined}, respectively. For each model's result, the originally proposed smoothing methods have been applied. For Dice, $^{\dagger}$ and $^{\ddagger}$ indicate significantly lower scores than Ours and Ours with mean filtering ($k=5$), respectively (under two-sided paired Wilcoxon signed-rank tests, all $p<0.001$, $n=44$).}
\label{tab:unsupervised_results}
\setlength{\tabcolsep}{6pt}
\renewcommand{\arraystretch}{1.3}
{
\footnotesize
\begin{tabular}{lccccc}
\toprule
\textbf{Method}
& \textbf{Dice $\uparrow$}
& \textbf{AP $\uparrow$}
& \textbf{HD $\downarrow$ (mm)}
& \textbf{MASD $\downarrow$ (mm)}
& \textbf{NSD $\uparrow$}
\\
\midrule

\makecell[l]{f-AnoGAN\\\citep{schlegl2019f}}
& \multicolumn{5}{c}{Unstable training and unable to achieve convergence.}
\\

\makecell[l]{DAE\\{\citep{kascenas2022denoising}}}
& $0.159 \pm 0.268^{\dagger\ddagger}$
& $\mathbf{0.665}$
& $30.38\pm15.41$
& $13.65\pm9.87$
& $0.176\pm0.210$
\\

\makecell[l]{AnoDDPM\\{\citep{wyatt2022anoddpm}}}
& \multicolumn{5}{c}{Unstable training with simplex noise.}
 \\

\makecell[l]{AutoDDPM\\{\citep{bercea2023mask}}}
& $0.085 \pm 0.126^{\dagger\ddagger}$
& $0.083$
& $31.96 \pm 15.59$
& $14.53 \pm 9.73$
& $0.169 \pm 0.194$
 \\

\makecell[l]{pDDPM\\{\citep{behrendt2024patched}}}
& $0.121 \pm 0.235^{\dagger\ddagger}$
& $0.084$
& $39.08 \pm 15.05$
& $21.16 \pm 11.08$
& $0.050 \pm 0.111$
 \\

\makecell[l]{THOR\\{\citep{bercea2024diffusion}}}
& $0.162 \pm 0.231^{\dagger\ddagger}$
& $0.281$
& $31.89 \pm 15.56$
& $14.35 \pm 9.68$
& $0.196 \pm 0.218$
 \\

\makecell[l]{IterMask$^2$\\{\citep{liang2024itermask}}}
& $0.121 \pm 0.235^{\dagger\ddagger}$
& $0.050$
& $38.87\pm 15.33$
& $21.09\pm 11.11$
& $0.050\pm 0.111$
 \\

\makecell[l]{MAD-AD\\{\citep{beizaee2025mad}}}
& $0.166 \pm 0.205^{\dagger\ddagger}$
& $0.492$
& $\underline{27.44 \pm 13.25}$
& $\underline{11.33 \pm 7.84}$
& $0.213 \pm 0.187$
 \\

\makecell[l]{DeCo-Diff\\{\citep{beizaee2025correcting}}}
& $0.131 \pm 0.214^{\dagger\ddagger}$
& $0.261$
& $30.79\pm 14.84$
& $14.86 \pm 9.51$
& $0.154 \pm 0.189$
 \\

\makecell[l]{ANDi\\{\citep{FROTSCHER2026103895}}}
& $0.108 \pm 0.199^{\dagger\ddagger}$
& $0.267$
& $ 33.09 \pm 14.79$
& $12.85 \pm 8.37$
& $0.139 \pm 0.179$
 \\

\makecell[l]{InvAD + U-Net\\{\citep{sakai2026invad}}}
& $ 0.131\pm 0.235^{\dagger\ddagger}$
& $0.174$
& $32.47\pm15.95$
& $16.26\pm10.02$
& $0.136\pm0.203$

 \\

\makecell[l]{InvAD  + original setup\\{\citep{sakai2026invad}}}
& $ 0.121\pm 0.235^{\dagger\ddagger}$
& $0.107$
& $39.09\pm15.05$
& $21.17\pm11.08$
& $0.050\pm0.1112$
\\

\midrule

\textbf{Ours}
& $\underline{0.232\pm0.239^{\ddagger}}$
& $	0.374$
& $29.29 \pm 16.43$
& $12.62 \pm 10.11$
& $\underline{0.284 \pm 0.256}$
 \\

\textbf{Ours + mean filtering ($k=5$)}
& $\mathbf{0.283\pm 0.244}$
& $\underline{0.505}$
& $\mathbf{25.45 \pm 15.63}$
& $\mathbf{9.40 \pm 8.49}$
& $\mathbf{0.355 \pm 0.236}$
 \\

\bottomrule
\end{tabular}%
}
\end{table*}
\begin{table*}[t]
\centering
\caption{
Performance of our proposed model stratified by subject-level lesion burden. Subjects were divided into small ($<1\%$), medium ($1\%$--$5\%$), and large ($\geq5\%$) lesion-burden groups according to the percentage of voxels labeled as lesions in each subject. MF5 denotes mean filtering with a kernel size of $5$ (rows shaded in gray). Results are reported as mean $\pm$ standard deviation where available, and the best result within each lesion-burden group is highlighted in bold.
}
\label{tab:lesion_burden_analysis}

{\footnotesize
\setlength{\tabcolsep}{5.5pt}
\renewcommand{\arraystretch}{1.3}

\begin{tabular}{llccccc}
\toprule
\textbf{Lesion burden}
& \textbf{Post-processing}
& \textbf{Dice $\uparrow$}
& \textbf{AP $\uparrow$}
& \textbf{HD $\downarrow$ (mm)}
& \textbf{MASD $\downarrow$ (mm)}
& \textbf{NSD $\uparrow$} \\
\midrule

\multirow{2}{*}{Small ($<1\%$), $n=22$}
& $--$
& $0.080 \pm 0.086$
& $0.030$
& $40.27 \pm 13.90$
& $17.25 \pm 9.58$
& $0.155 \pm 0.111$ \\

& \cellcolor{gray!12} MF5
& \cellcolor{gray!12} $\mathbf{0.141 \pm 0.154}$
& \cellcolor{gray!12} $\mathbf{0.135}$
& \cellcolor{gray!12} $\mathbf{33.60 \pm 15.19}$
& \cellcolor{gray!12} $\mathbf{10.49 \pm 5.64}$
& \cellcolor{gray!12} $\mathbf{0.262 \pm 0.208}$ \\

\addlinespace[3pt]

\multirow{2}{*}{Medium ($1\%$--$5\%$), $n=12$}
& $--$
& $0.287 \pm 0.148$
& $0.186$
& $26.39 \pm 8.77$
& $7.80 \pm 3.53$
& $0.388 \pm 0.150$ \\

& \cellcolor{gray!12} MF5
& \cellcolor{gray!12} $\mathbf{0.377 \pm 0.208}$
& \cellcolor{gray!12} $\mathbf{0.356}$
& \cellcolor{gray!12} $\mathbf{18.43 \pm 6.92}$
& \cellcolor{gray!12} $\mathbf{5.47 \pm 3.54}$
& \cellcolor{gray!12} $\mathbf{0.487 \pm 0.232}$ \\

\addlinespace[3pt]

\multirow{2}{*}{Large ($\geq5\%$), $n=10$}
& $--$
& $\mathbf{0.603 \pm 0.147}$
& 0.541
& $10.23 \pm 5.03$
& $\mathbf{2.68 \pm 1.59}$
& $\mathbf{0.640 \pm 0.160}$ \\

& \cellcolor{gray!12} MF5
& \cellcolor{gray!12} $0.600 \pm 0.167$
& \cellcolor{gray!12} $\mathbf{0.562}$
& \cellcolor{gray!12} $\mathbf{10.13 \pm 4.75}$
& \cellcolor{gray!12} $2.76 \pm 0.93$
& \cellcolor{gray!12} $0.582 \pm 0.137$ \\

\bottomrule
\end{tabular}
}
\end{table*}

Table~\ref{tab:unsupervised_results} compares the proposed method with existing state-of-the-arts unsupervised anomaly detection approaches. The compared approaches were trained over the same training subjects and used their original proposed post-processing smooth methods. Among the evaluated baselines, f-AnoGAN \citep{schlegl2019f} failed to converge reliably and AnoDDPM \citep{wyatt2022anoddpm} remained unstable when trained with Simplex noise. Therefore, quantitative results are not reported for these two methods.

Without spatial filtering, our method achieved a Dice score of \(0.232\), significantly exceeding the best-performing converged baselines, MAD-AD \citep{beizaee2025mad} with Dice $0.166$ ($p<0.001$, $n= 44$). It also obtained an NSD of \(0.284\), compared with the best baseline MAD-AD's NSD of \(0.213\). The corresponding HD and MASD were \(29.29\) mm and \(12.62\) mm, respectively, representing the second-best performance for both distance-based metrics after MAD-AD. The unfiltered anomaly map achieved an AP of 0.374, which was lower than those of DAE (0.665) and MAD-AD (0.492), but higher than the remaining diffusion-based and feature-based baselines.

Applying mean filtering with a kernel size of \(k=5\) consistently improved the proposed method across all reported metrics. Dice significantly increased from \(0.232\) to \(0.283\) (absolute gain of $0.051$; $p=0.001$, $n= 44$). AP increased from 0.374 to 0.505, while HD decreased from \(29.29\) to \(25.45\) mm, and MASD decreased from \(12.62\) to \(9.40\) mm. NSD also increased from \(0.284\) to \(0.355\). The mean-filtered model therefore achieved the best Dice, HD, MASD, and NSD among all successfully trained methods.

Relative to MAD-AD, the strongest competing baseline for spatial localization, the mean-filtered model significantly improved Dice by 0.117 ($p<0.001$, $n= 44$), reduced HD by 1.99 mm, reduced MASD by 1.93 mm, and increased NSD by 0.142. Although DAE \citep{kascenas2022denoising} achieved the highest AP of 0.665, its Dice $0.159$ and NSD $0.176$ were substantially lower than those of our proposed mean-filtered method. Overall, these findings indicate that the proposed approach provides more accurate lesion overlap and boundary localization, while DAE retains an advantage in global voxel-ranking performance.

To assess the effect of lesion burden on detection performance, we further stratified the test subjects according to their ground-truth lesion percentage into small ($<1\%$), medium ($1\%$--$5\%$), and large ($\geq5\%$) lesion-burden groups, as following \cite{rao2025spatial} in Table~\ref{tab:lesion_burden_analysis}. Before mean filtering, performance improved consistently with increasing lesion burden: Dice increased from $0.080$ for the small-lesion group to $0.287$ and $0.603$ for
the medium- and large-lesion groups, respectively. Correspondingly, HD decreased from $40.27$~mm to $26.39$~mm and $10.23$~mm, while NSD increased from $0.155$ to $0.388$ and $0.640$.

Mean filtering provided the greatest benefit for subjects with small and medium lesion burdens (Table~\ref{tab:lesion_burden_analysis}, gray-shaded rows). For the small-lesion group, mean filtering with kernel size of $5$ (MF5) increased Dice from $0.080$ to $0.141$, AP from $0.030$ to $0.135$, and NSD from $0.155$ to $0.262$. For the medium-lesion group, Dice increased from $0.287$ to $0.377$, AP from $0.186$ to $0.356$, and NSD from $0.388$ to $0.487$. In contrast, performance for the large-lesion group was already comparatively high before filtering, and MF5 produced only a small increase in AP ($0.541$ to $0.562$) and HD improvement ($10.23$ to $10.13$~mm), while Dice remained nearly unchanged ($0.603$ versus $0.600$).

Fig.~\ref{fig:anomaly_demos} compares anomaly maps from selected top-performing baselines with those produced by our method in subjects
with different lesion burdens. In the illustrated cases, DAE \citep{kascenas2022denoising} produces spatially extensive responses beyond the annotated lesions, whereas
ANDi \citep{FROTSCHER2026103895} exhibits scattered responses with limited separation between
lesion and non-lesion tissue. MAD-AD \citep{beizaee2025mad} produces more localized
responses but also highlights regions outside the lesion annotations, particularly in the small-lesion example. Our method shows closer spatial correspondence with the annotated lesions in the medium-
and large-burden examples, although scattered responses remain outside the lesion boundaries. Mean filtering reduces isolated responses and produces smoother anomaly maps while preserving the lesion associated responses. The small-lesion example remains challenging, where the lesion is detected but is less noticeable with surrounding false positive responses. Some apparent false positive responses occur in regions with relatively low ADC values or interhemispheric asymmetry in ADC intensity, indicating that the directional constraint does not fully suppress responses outside the annotated lesions.

\subsection{Comparison with Supervised ADC-Only Models}

\begin{table*}[t]
\centering
\caption{
Effect of incorporating population-derived ADC \(Z\)-map information. ADC-only models serve as the reference configurations (rows shaded in gray). For the supervised models, the \(Z\)-map and a thresholded binary \(Z\)-mask (\(Z\le-2.0\)) are provided as additional input channels \citep{almansour2025enhancing,bao2025bonbid,rao2025spatial}. For the population-derived statistical baseline, \(-Z\) is used as the continuous anomaly score for AP, whereas \(Z\le-2.0\) is used to obtain the binary mask for Dice and surface-distance metrics. Our unsupervised method uses the \(Z\)-map only during post processing. AP is not reported for this post-processing result because it directly produces a binary lesion mask. The best result within each model is shown in \textbf{bold}. For Dice, $^{\dagger}$ indicates a significantly lower score than our proposed method with $Z$-map post-processing (under two-sided paired Wilcoxon signed-rank test at significance level $\alpha=0.05$, $n=44$).}
\label{tab:supervised_result}
\setlength{\tabcolsep}{5.5pt}
\renewcommand{\arraystretch}{1.3}

{\footnotesize
\begin{tabular}{llccccc}
\toprule
\textbf{Model}
& \textbf{Input or refinement}
& \textbf{Dice $\uparrow$}
& \textbf{AP $\uparrow$}
& \textbf{HD $\downarrow$ (mm)}
& \textbf{MASD $\downarrow$ (mm)}
& \textbf{NSD $\uparrow$} \\
\midrule

\multicolumn{7}{l}{\textit{Supervised segmentation methods}} \\[2pt]

\multirow{3}{*}{\makecell[l]{Swin-UNETR\\ \citep{almansour2025enhancing}}}
& \cellcolor{gray!12} ADC only
& \cellcolor{gray!12} $0.311 \pm 0.251^{\dagger}$
& \cellcolor{gray!12}$0.645$
& \cellcolor{gray!12}$22.56 \pm 13.39$
& \cellcolor{gray!12}$8.31 \pm 8.38$
& \cellcolor{gray!12}$0.406 \pm 0.247$ \\

& ADC + $Z$-map
& $0.553 \pm 0.260^{\dagger}$
& $0.918$
& $\mathbf{16.07 \pm 12.00}$
& $4.10 \pm 6.31$
& $0.674 \pm 0.256$ \\

& ADC + $Z$-map + $Z \le -2.0$
& $\mathbf{0.657 \pm 0.258}$
& $\mathbf{0.927}$
& $18.32 \pm 15.57$
& $\mathbf{3.89 \pm 5.46}$
& $\mathbf{0.724 \pm 0.242}$ \\

\addlinespace[3pt]

\multirow{3}{*}{\makecell[l]{UNet++\\\citep{rao2025spatial}}}
& \cellcolor{gray!12} ADC only
& \cellcolor{gray!12} $0.237 \pm 0.260^{\dagger}$
& \cellcolor{gray!12}$0.803$
& \cellcolor{gray!12}$23.58 \pm 12.80$
& \cellcolor{gray!12}$9.68 \pm 8.37$
& \cellcolor{gray!12}$0.298 \pm 0.235$ \\

& ADC + $Z$-map
& $0.578 \pm 0.246$
& $\mathbf{0.915}$
& $\mathbf{15.88 \pm 12.77}$
& $3.84 \pm 5.55$
& $0.694 \pm 0.249$ \\

& ADC + $Z$-map + $Z \le -2.0$
& $\mathbf{0.609 \pm 0.250}$
& $0.913$
& $17.51 \pm 14.47$
& $\mathbf{3.71 \pm 4.71}$
& $\mathbf{0.700 \pm 0.241}$ \\

\midrule

\multicolumn{7}{l}{
    \textit{Population-derived statistical reference}
} \\[2pt]

\makecell[l]{Population ADC\\\(Z\)-map \citep{bao2025bonbid}}
& \makecell[l]{\(Z\)-map only\\(\(Z\le-2.0\) for segmentation)}
& $0.583 \pm 0.268^{\dagger}$
& $0.922$
& $19.40 \pm 13.45$
& $4.72 \pm 6.17$
& $0.641 \pm 0.256$ \\

\midrule

\multicolumn{7}{l}{\textit{Unsupervised anomaly detection method}} \\[2pt]

\multirow{2}{*}{\textbf{Ours}}
& \cellcolor{gray!12} ADC only
& \cellcolor{gray!12}$0.232 \pm 0.239^{\dagger}$
& \cellcolor{gray!12}$0.374$
& \cellcolor{gray!12}$29.29 \pm 16.43$
& \cellcolor{gray!12} $12.62 \pm 10.11$
& \cellcolor{gray!12}$0.284 \pm  0.256$ \\

& ADC + $Z$-map post-processing
& $\mathbf{0.602 \pm 0.269}$
& $--$
& $\mathbf{19.11 \pm 19.00}$
& $\mathbf{4.73 \pm 6.40}$
& $\mathbf{0.667 \pm 0.262}$ \\

\bottomrule
\end{tabular}%
}
\end{table*}

Table~\ref{tab:supervised_result} compares the proposed unsupervised anomaly detection method with supervised Swin-UNETR \citep{almansour2025enhancing} and UNet++ \citep{rao2025spatial} segmentation models under matched ADC-only and population \(Z\)-map enhanced settings, which are the top performed models from the BONBID-HIE challenge \citep{bao2025bonbid}. In the ADC-only setting (the gray-shaded rows in Table~\ref{tab:supervised_result}), our method achieved a Dice score of \(0.232\), which was comparable to the supervised UNet++ model of $0.237$, and an NSD of \(0.284\), approaching the UNet++ NSD of \(0.298\). Swin-UNETR achieved highest ADC-only Dice and NSD values of \(0.311\) and \(0.406\), respectively.

Despite the comparable overlap with UNet++, the ADC-only anomaly detection model remained less competitive in voxel-ranking and boundary-distance performance. It achieved an AP of 0.374, compared with 0.645 for Swin-UNETR and 0.803 for UNet++. Similarly, its HD of \(29.29\) mm and MASD of \(12.62\) mm were higher than those of both supervised models. These findings show that the proposed method can approach a supervised segmentation network in terms of lesion overlap and tolerance-based surface agreement without being optimized using a voxel-wise segmentation loss, although a gap remains in anomaly score ranking and precise boundary localization.

\subsection{Effect of Combining Population-Derived ADC \(Z\)-Map Information}

We additionally evaluated an optional post-processing procedure incorporating a population-derived ADC $Z$-map (Table~\ref{tab:supervised_result}). For the supervised models, the continuous $Z$-map and, a thresholded $Z\le-2.0$ mask \citep{rao2025spatial} were supplied as additional input channels during training and inference. In contrast, our diffusion model remained ADC-only, and the $Z$-map was used exclusively during post-processing to identify a high-confidence low-ADC core and to retain nearby diffusion-supported components. We also evaluated a $Z$-map-only baseline using $Z\le-2.0$ for binary segmentation.

The $Z$-map-only baseline achieved a Dice score of $0.583$, indicating that population-derived ADC information alone provides strong lesion localization performance. Incorporating our diffusion-derived anomaly information increased Dice to $0.602$ and NSD from $0.641$ to $0.667$, corresponding to absolute gains of $0.019$ and $0.026$, respectively. HD decreased slightly from $19.40$ to $19.11$~mm, whereas MASD remained essentially unchanged ($4.72$ versus $4.73$~mm). These results demonstrate a modest but statistically significant improvement in Dice ($p=0.0020$, $n=44$), accompanied by a numerical increase in surface agreement over the $Z$-map-only baseline.

Relative to the ADC-only diffusion prediction, $Z$-map guided
post-processing increased Dice from $0.232$ to $0.602$ (absolute gain of $0.370$; $p<0.001, n=44$) and NSD from $0.284$ to $0.667$. The post-processing refined prediction achieved significantly higher Dice than Swin-UNETR using ADC and the continuous $Z$-map ($0.602$ versus $0.553$, absolute gain of $0.049$; $p=0.0020, n=44$). Its Dice was also numerically higher than that of UNet++ using ADC and the continuous $Z$-map ($0.578$), although this difference was not statistically significant ($p=0.156, n=44$). The post-processing refined Dice remained numerically below those of Swin-UNETR and UNet++ additionally supplied with the thresholded binary $Z$-mask (Dice of $0.657$ and $0.609$, respectively). Its NSD of $0.667$ was slightly below the range achieved by the $Z$-map assisted supervised models ($0.674$--$0.724$). AP was not reported for our post-processing procedure because it directly produces a binary lesion mask.

Fig.~\ref{fig:ablation_demos} illustrates the effect of
post-processing on the ADC-only diffusion prediction
in a representative case. Many spatially isolated
responses are removed, while the main lesion-associated
regions are retained. The resulting segmentation
broadly follows regions with low ADC $Z$-values and
overlaps with the expert annotations outlined in cyan.

\subsection{Ablation Analysis of Framework Components}
\label{subsec:ablation_component}

\begin{table*}[t]
\centering
\caption{
Ablation study of the proposed framework. All compared models applied the HIE pathology informed directional information. Results are reported as mean $\pm$ standard deviation where applicable.  The best and second-best
results are shown in \textbf{bold} and \underline{underlined},
respectively.
}
\label{tab:component_ablation}
\footnotesize
\setlength{\tabcolsep}{7pt}
\renewcommand{\arraystretch}{1.3}
\begin{tabular}{lccccc}
\toprule
\textbf{Configuration}
& \textbf{Dice $\uparrow$}
& \textbf{AP $\uparrow$}
& \textbf{HD $\downarrow$ (mm)}
& \textbf{MASD $\downarrow$ (mm)}
& \textbf{NSD $\uparrow$} \\
\midrule

DDPM
& $0.167\pm0.236$ 
& $0.235$
&  $32.03\pm16.62$ 
&  $14.58\pm10.03$ 
&  $0.212\pm0.242$ \\

\addlinespace[2pt]
\makecell[l]{\quad $+$ ControlNet \\ \quad (healthy-restoration, CFG=0.8)}
&  $0.121\pm0.190$ 
& \underline{$0.414$}
& $39.44\pm16.03$
& $17.47 \pm10.19$
& $0.169 \pm0.193$ \\

\addlinespace[2pt]
\makecell[l]{\quad $+$ ControlNet \\ \quad (lesion-emphasizing, CFG=7.5)}
& $0.138\pm0.242$ 
& $0.332$
& $31.30\pm16.29$ 
& $15.31\pm10.37$ 
& $0.160 \pm0.222$ \\

\addlinespace[2pt]

\makecell[l]{\quad $+$ Multi noise-level \\ \quad discrepancy accumulation}
& \underline{$0.232\pm0.239$}
& $0.374$
&\underline{$29.29\pm16.43$}
&  \underline{$12.62\pm10.11$}
&  \underline{$0.284 \pm0.256$} \\

\addlinespace[2pt]
\quad $+$ Mean filtering ($k=5$)
& $\mathbf{0.283\pm0.244}$
& $\mathbf{0.505}$
& $\mathbf{25.45\pm15.63}$ 
& $\mathbf{9.40\pm8.49}$
& $\mathbf{0.355\pm0.236}$ \\

\bottomrule
\end{tabular}
\end{table*}

Table~\ref{tab:component_ablation} evaluates the contributions of the principal components of the proposed framework. The conventional DDPM baseline achieved a Dice score of $0.167 $, an AP of $0.235$, and an NSD of $0.212 $, with an HD of $32.03$ mm and an MASD of $14.58$ mm.

Introducing subject-specific conditioning through ControlNet and deriving the anomaly map from a single CFG-based reconstruction (described in Sec.~\ref{sec:cfg_interpretation}) did not result in consistent improvements across the evaluation metrics. With a lesion-recovery setting $w=0.8$, AP increased from $0.235$ to $0.414$. However, Dice, HD, MASD, and NSD deteriorated. This result indicates that reducing the contribution of the subject condition can increase voxel-wise sensitivity to abnormal regions, while simultaneously introducing reconstruction differences that are spatially inconsistent with the lesion boundaries. Increasing the guidance scale to $w=7.5$, corresponding to the lesion-strengthening setting, partially improved Dice and the surface-distance metrics relative to $w=0.8$, but reduced AP to $0.332$ and did not provide a consistent advantage over the DDPM baseline. These findings demonstrate the limited and inconsistent lesion sensitivity of anomaly maps derived from a single reconstructed image. 

The proposed multi noise-level discrepancy accumulation directly compares the conditional and unconditional noise predictions and aggregates their directionally consistent differences across the selected reverse diffusion steps. This component significantly improved Dice from \(0.167\) to \(0.232\) (absolute gain of $0.065$; $p< 0.001$, $n = 44$), reduced HD from \(32.03\) to \(29.29\) mm, and reduced MASD from \(14.58\) to \(12.62\) mm. NSD also increased from \(0.212\) to \(0.284\). Although its AP of \(0.374\) was lower than that obtained with the \(w=0.8\) reconstruction, the consistently improved overlap and surface-distance metrics indicate more accurate spatial localization of the detected abnormalities. Importantly, because the discrepancy is computed directly from
\(\boldsymbol{\epsilon}^{\mathrm{cond}}-\boldsymbol{\epsilon}^{\mathrm{uncond}}\), its magnitude is independent of the CFG scale \(w\).

Finally, applying mean filtering with a kernel size of \(k=5\) produced the best performance across all metrics. This improvement suggests that spatial smoothing suppresses isolated high-frequency responses while preserving spatially coherent lesion evidence.

Fig.~\ref{fig:ablation_demos} illustrates the differences
between reconstruction-based anomaly maps and our discrepancy-based
approach in a representative case. DDPM alone produces substantial anatomical deviations from the input image, whereas ControlNet conditioning better preserves
subject-specific anatomy. However, both DDPM and ControlNet-guided reconstruction with CFG $=0.8$ produce weak anomaly responses within the annotated lesion region. Increasing the guidance scale to CFG $=7.5$ produces higher
anomaly scores overall, making the lesion region more noticeable. However, elevated scores also occur in surrounding non-lesion tissue, limiting spatial specificity. In contrast, our method produces more spatially confined responses
that follow the annotated lesion distribution, although some
lesion regions remain weakly highlighted and scattered responses persist outside the annotation. Mean filtering reduces isolated responses and produces a smoother map while retaining the main lesion-associated pattern. These observations complement the
quantitative ablation results and illustrate the benefit of
aggregating conditional--unconditional discrepancies across
noise levels rather than relying on a final reconstruction residual.

\subsection{Ablation Analysis of Mean Filtering Kernel Size}

\begin{table*}[t]
\centering
\caption{
Ablation study of the mean-filter kernel size applied to the  direction-aware anomaly map. Kernel size $k=0$ indicates that no mean filtering was applied. The gray row denotes the kernel size used in the main experiments. Results are reported as mean $\pm$ standard deviation where applicable, and the best value for each metric is highlighted in \textbf{bold}.
}
\label{tab:mean_filter_ablation}
\footnotesize
\setlength{\tabcolsep}{7pt}
\renewcommand{\arraystretch}{1.3}

\begin{tabular}{cccccc}
\toprule
\textbf{Kernel size $k$}
& \textbf{Dice $\uparrow$}
& \textbf{AP $\uparrow$}
& \textbf{HD $\downarrow$ (mm)}
& \textbf{MASD $\downarrow$ (mm)}
& \textbf{NSD $\uparrow$} \\
\midrule

0
& $0.232 \pm 0.239$
& $0.374$
& $29.29 \pm 16.43$
& $12.62 \pm 10.11$ 
& $0.284 \pm 0.256$ \\

3
& $0.262 \pm 0.245$
& $0.481$
& $27.24 \pm 16.37$
& $11.01 \pm 9.74$
& $0.324 \pm 0.250$ \\

\rowcolor{gray!12}
5
& $0.283 \pm 0.244$
& $\mathbf{0.505}$
& $\mathbf{25.45\pm15.63}$
&  $\mathbf{9.40\pm8.49}$ 
&  $\mathbf{0.355\pm0.236}$  \\

7
&   $\mathbf{0.287\pm0.252}$ 
& $0.503$
& $26.20 \pm 15.47$
& $9.58 \pm 8.43$
& $0.348 \pm 0.229$ \\

8
& $0.282  \pm 0.255$
& 	$0.497$
& $26.51  \pm 15.14$
& $9.74\pm 8.38$
& $0.331  \pm 0.224$ \\

10
& $0.270  \pm 0.259$
& $0.485$
& $28.16 \pm16.76$
& $10.97  \pm 9.52$
& $0.304  \pm 0.222$\\

\bottomrule
\end{tabular}
\end{table*}

Table~\ref{tab:mean_filter_ablation} evaluates the effect of the mean filtering kernel size on the direction-aware anomaly map. Here, \(k=0\) denotes the unfiltered anomaly map. Applying a moderate degree of spatial smoothing consistently improved performance over the unfiltered result. Increasing the kernel size from \(k=0\) to \(k=3\) improved Dice from \(0.232\) to \(0.262\), AP from \(0.374\) to \(0.481\) and NSD from $0.284$ to $0.324$, while reducing HD from \(29.29\) to \(27.24\) mm and MASD from \(12.62\) to \(11.01\) mm. Further increasing the kernel size to \(k=5\) yielded the highest AP (\(0.505\)), the lowest HD (\(25.45\) mm), the lowest MASD (\(9.40\) mm), and the highest NSD (\(0.355\)). These improvements indicate that mean filtering effectively suppresses spatially isolated responses while preserving coherent lesion evidence.

A kernel size of \(k=7\) produced the numerically highest Dice score of \(0.287\), compared with \(0.283\) for \(k=5\), but resulted in slightly inferior AP and surface-distance based metrics. Performance subsequently declined with larger kernels. At \(k=10\), Dice decreased to \(0.270\), HD increased to \(28.16\) mm, MASD increased to $10.97$ mm and NSD decreased to \(0.304\). This degradation suggests that excessive smoothing increasingly mixes lesion and healthy-tissue responses, reducing boundary accuracy and the localization of smaller lesions. Thus, the validation-selected kernel size of \(k=5\) also provided a favorable balance of overlap, ranking, and surface agreement on the test set.

\subsection{Ablation Analysis of Discrepancy Formulations}

\begin{table*}[t]
\centering
\caption{Ablation of the anomaly-score representation compared with sequential reverse propagation and discrepancy to forward process noise. ``Independent'' indicates that each noise level was sampled directly from the original patient ADC image without propagating a reverse-diffusion chain, and ``Reverse chain'' evaluates successive states along a sequential reverse-diffusion trajectory (Eq.~\eqref{eq:chained_discrepancy}). The final row replaces the learned unconditional noise prediction with the Gaussian noise $\epsilon_t$ sampled during forward corruption (Eq.~\eqref{eq:forward_noise_discrepancy}). The biological directional constraint was enabled for all configurations. Rows shaded in gray use a mean filtering with kernel size of $5$ (MF5). Values are reported as mean $\pm$ standard deviation where available, and the best result for each metric is highlighted in \textbf{bold}.}
\label{tab:score_chain_ablation}

{
\footnotesize
\renewcommand{\arraystretch}{1.3}
\begin{tabular}{@{}lcccccc@{}}
\toprule
\textbf{Score representation}
& \textbf{Evaluation}
& \textbf{Dice $\uparrow$}
& \textbf{AP $\uparrow$}
& \textbf{HD $\downarrow$ (mm)}
& \textbf{MASD $\downarrow$ (mm)}
& \textbf{NSD $\uparrow$} \\
\midrule
$\epsilon_t^{\text{cond}}-\epsilon_t^{\text{uncond}}$
& Independent
& $0.232\pm 0.239$
& $0.374$
& $29.29 \pm 16.43$
& $12.62 \pm 10.11$
& $0.284 \pm 0.256$ \\

\rowcolor{gray!15}
$\epsilon_t^{\text{cond}}-\epsilon_t^{\text{uncond}}$ $+$ MF5
& Independent
&  $\mathbf{0.283\pm 0.244}$
& $\mathbf{0.505}$
&   $\mathbf{25.45\pm15.63}$
&  $\mathbf{9.40\pm  8.49}$
&  $\mathbf{0.355\pm 0.236}$ \\

\addlinespace[2pt]

$\epsilon_t^{\text{cond}}-\epsilon_t^{\text{uncond}}$
& Reverse chain
& $0.070  \pm 0.119$
& $0.031$
& $32.05  \pm 16.01$
& $15.85  \pm 10.36$
& $0.170  \pm 0.238$ \\

\rowcolor{gray!15}
$\epsilon_t^{\text{cond}}-\epsilon_t^{\text{uncond}}$ $+$ MF5
& Reverse chain
& $0.070  \pm 0.108$
& $0.036$
& $49.59  \pm 25.26$
& $23.36  \pm 15.44$
& $0.114  \pm 0.154$ \\

\makecell{$\epsilon_t^{\text{cond}}-\epsilon_t$ \\using forward noise}
& Independent
& $0.045  \pm 0.211$
& $0.017$
& $75.30  \pm 9.29$
& $52.69  \pm 7.76$
& $0.000  \pm 0.000$ \\
\addlinespace[2pt]
\bottomrule
\end{tabular}%
}
\end{table*}

Table~\ref{tab:score_chain_ablation} compares independent noise-level evaluation with reverse-chain accumulation and a formulation that replaces the unconditional prediction with the sampled forward noise, as described in Sec. \ref{subsec:discrepancy_formulation_ablation}. The pathology-informed directional constraint was applied to all configurations. 

Without mean filtering, independent evaluation outperformed reverse-chain accumulation (Eq.~\eqref{eq:chained_discrepancy}) in Dice ($0.232$ versus $0.070$), AP ($0.374$ versus $0.031$), and all surface metrics. This advantage persisted after mean filtering: reverse-chain formulation's Dice remained at $0.070$ and AP increased only to $0.036$, while HD and MASD worsened to $49.59$ and $23.36$~mm, respectively, and NSD decreased to $0.114$.
The reduced performance of chain-dependent accumulation suggests that maintaining a continuous reverse trajectory is not advantageous for discrepancy-based anomaly localization. Because each state is generated from the preceding reverse estimate, stochastic perturbations and reconstruction errors introduced at earlier steps propagate through the remainder of the trajectory. Moreover, as the reverse process progressively modifies the image, subtle pathological information present in the original subject image may be attenuated, causing later discrepancy estimates to become less directly associated with the observed abnormality. In contrast, independently constructing the input at each noise level connects every comparison to the same observed image and prevents errors at one timestep from affecting discrepancy measurements at subsequent timesteps.

Replacing the unconditional prediction with the sampled forward noise (Eq.~\eqref{eq:forward_noise_discrepancy}) produced the lowest performance across all metrics, with a Dice score of $0.045$, AP of $0.017$, HD of $75.30$~mm, MASD of $52.69$~mm, and NSD of $0.000$. The sampled forward noise is independent of the subject anatomy and does not encode the learned healthy-data distribution. Consequently, its difference from the conditional prediction primarily reflects noise-prediction error rather than deviation from a healthy reference. By comparison, the unconditional branch provides the denoising response favored by the learned healthy prior for the same noisy input.

The results therefore support both central components of the proposed formulation: independent evaluation across noise levels prevents trajectory-dependent error propagation, while the learned unconditional prediction provides a meaningful healthy prior reference for detecting subject-specific pathological deviations.

\subsection{Verification of the HIE Pathology Directional Prior}

\begin{table*}[t]
\centering
\caption{
Directional characteristics of the discrepancy across the selected noise levels. The discrepancy is defined as $d_t(\mathbf{v}) =
\epsilon_t^{\mathrm{cond}}(\mathbf{v})-\epsilon_t^{\mathrm{uncond}}(\mathbf{v})$. A positive value indicates that the healthy-prior response predicts a higher ADC value than the patient-conditioned response, consistent with the expected direction of reduced ADC in acute HIE injury. Values are reported as the timestep-wise mean and 95\% confidence interval across the selected noise
levels ($n=|\mathcal{T}|=125$).
}
\label{tab:directional_discrepancy_stat}

{\footnotesize
\setlength{\tabcolsep}{6pt}
\renewcommand{\arraystretch}{1.3}

\begin{tabular}{@{}lccc@{}}
\toprule
\textbf{Statistic}
& \makecell{\textbf{Lesion}\\\textbf{Mean (95\% CI)}}
& \makecell{\textbf{Background}\\\textbf{Mean (95\% CI)}}
\\
\midrule

\makecell[l]{Positive discrepancy probability, $\Pr(d_t>0)$}
&
\makecell{$0.886$\\$(0.884,\,0.888)$}
&
\makecell{$0.530$\\$(0.526,\,0.533)$}

\\[2pt]

\makecell[l]{Mean positive discrepancy,
$\mathbb{E}[d_t\mid d_t>0]$}
&
\makecell{$0.266$\\$(0.260,\, 0.272)$}
&
\makecell{$0.155$\\$( 0.151,\,0.159)$}
&
\\[2pt]

\makecell[l]{Mean negative discrepancy,
$\mathbb{E}[d_t\mid d_t<0]$}
&
\makecell{$-0.107$\\$(-0.110,\, -0.104)$}
&
\makecell{$-0.176$\\$( -0.178,\,-0.174)$} 
\\[2pt]

\makecell[l]{Mean discrepancy magnitude, $\mathbb{E}[|d_t|]$}
&
\makecell{$0.248$\\$(0.242,\,0.254)$}
&
\makecell{$0.165$\\$(0.162,\,0.168)$}
\\

\bottomrule
\end{tabular}
}
\end{table*}

{\small
\begin{table*}[t]
\centering
\caption{
Ablation of the pathology-informed directional sign constraint and mean-filter post-processing. MF5 denotes mean filtering with a kernel size of 5. Rows using MF5 are shaded, and the best result for each metric is shown in \textbf{bold}.
}
\label{tab:pathology_sign_ablation}
\footnotesize
\setlength{\tabcolsep}{6pt}
\renewcommand{\arraystretch}{1.3}

\begin{tabular}{lccccccc}
\toprule
\makecell{\textbf{HIE Pathology prior}\\\textbf{directional sign $\pm$}}
& \textbf{Post-processing}
& \textbf{Dice $\uparrow$}
& \textbf{AP $\uparrow$}
& \textbf{HD $\downarrow$ (mm)}
& \textbf{MASD $\downarrow$ (mm)}
& \textbf{NSD $\uparrow$} \\
\midrule

No
& $--$
& $0.115 \pm 0.148$
& $0.222$
& $33.99 \pm 16.24$
& $15.18 \pm 10.04$
& $0.189 \pm 0.184$ \\

\rowcolor{gray!15}
No
& MF5
& $0.209 \pm 0.221$
& $0.490$
& $30.16 \pm 14.83$
& $11.48 \pm 8.82$
& $0.266 \pm 0.207$ \\

\midrule

Yes
& $--$
& $0.232 \pm 0.239$
& $0.374$
& $29.29 \pm 16.43$
& $12.62 \pm 10.11$
& $0.284 \pm 0.256$ \\

\rowcolor{gray!15}
Yes
& MF5
& $\mathbf{0.283 \pm 0.244}$
& $\mathbf{0.505}$
& $\mathbf{25.45 \pm 15.63}$
& $\mathbf{9.40 \pm 8.49}$
& $\mathbf{0.355 \pm 0.236}$ \\

\bottomrule
\end{tabular}
\end{table*}}
{\small\begin{table*}[t]
\centering
\caption{Quantitative comparison of unsupervised anomaly-detection baselines with introducing HIE pathology directional prior information at final reconstruction. Values are reported as mean $\pm$ standard deviation where available. The best result for each metric is highlighted in \textbf{bold}. For each model's result, the originally proposed smoothing methods have been applied. For Dice, $^{\dagger}$ and $^{\ddagger}$ indicate significantly lower scores than Ours and Ours with mean filtering ($k=5$), respectively (under two-sided paired Wilcoxon signed-rank tests, all $p<0.001$, $n=44$).}
\label{tab:unsupervised_bio_prior}
{
\footnotesize
\setlength{\tabcolsep}{7pt}
\renewcommand{\arraystretch}{1.3}
\begin{tabular}{@{}llccccc@{}}
\toprule
\textbf{Model}
& \textbf{Dice $\uparrow$}
& \textbf{AP $\uparrow$}
& \textbf{HD $\downarrow$ (mm)} 
& \textbf{MASD $\downarrow$ (mm)}
& \textbf{NSD $\uparrow$} \\
\midrule

DAE \citep{kascenas2022denoising}
& $0.172  \pm 0.272^{\dagger\ddagger}$
& $\mathbf{0.680}$
&  $\mathbf{25.13\pm 13.11}$
& $13.47  \pm 9.56$
& $0.188  \pm 0.219$\\

AutoDDPM \citep{bercea2023mask}
& $0.104 \pm 0.135^{\dagger\ddagger}$
& $0.119$
&$30.78  \pm 15.99$
& $13.12  \pm 9.59$
& $0.209  \pm 0.206$ \\

pDDPM \citep{behrendt2024patched}
& $0.121  \pm 0.235^{\dagger\ddagger}$
& $0.084$
& $39.07  \pm 15.07$
& $21.17  \pm 11.08$
& $0.050  \pm 0.111$ \\

MAD-AD \citep{beizaee2025mad}
&   $\mathbf{0.173\pm 0.198}^{\dagger\ddagger}$
& $0.491$
& $27.44  \pm 13.67$
&  $\mathbf{11.16\pm 7.99}$
&  $\mathbf{0.226\pm 0.192}$ \\

\bottomrule
\end{tabular}%
}
\end{table*}}

We first examined whether the direction of the conditional--unconditional discrepancy was consistent with the expected reduction in ADC associated with acute HIE lesions. Table~\ref{tab:directional_discrepancy_stat} summarizes the discrepancy
\(d_t(i)=\epsilon^{\mathrm{cond}}_{t,i}-\epsilon^{\mathrm{uncond}}_{t,i}\)
within lesion and background voxels across the selected noise levels. Through the positive scaling relationship established in the Methods Sec. \ref{subsec:discrepancy_score_accum}, positive value is equivalent to the healthy-prior response predicting a higher ADC value than the subject-conditioned response. Lesion voxels exhibited a positive discrepancy at \(88.6\%\) of the evaluated voxel--timestep pairs, compared with \(53.0\%\) in the background. The mean positive discrepancy was 0.266 in lesion versus 0.155 in background and the mean negative discrepancy was -0.107 and $-0.176$ respectively. These observations indicate that lesion voxels exhibit not only larger conditional--unconditional differences but also a systematic shift in the direction expected for reduced ADC. In contrast, background discrepancies nearly symmetrically occur in both directions.

Table~\ref{tab:pathology_sign_ablation} further evaluates whether incorporating this directional prior improves anomaly localization with and without mean filtering. Without mean filtering, the directional constraint significantly increased Dice from $0.115$ to $0.232$ (absolute gain of $0.117$; $p<0.001$, $n=44$). AP increased from $0.222$ to $0.374$, while HD decreased from $33.99$ to $29.29$~mm, MASD decreased from $15.18$ to $12.62$~mm, and NSD increased from $0.189$ to $0.284$. The improvement persisted when both formulations used mean filtering with a kernel size of $5$. Adding the directional constraint increased Dice from $0.209$ to $0.283$ (absolute gain of $0.074$; $p<0.001$, $n=44$). HD decreased from $30.16$ to $25.45$~mm, MASD decreased from $11.48$ to $9.40$~mm, and NSD increased from $0.266$ to $0.355$, alongside a smaller increase in AP from $0.490$ to $0.505$. Thus, the directional constraint improved all evaluated metrics both with and without mean filtering.

To evaluate whether the low-ADC directional prior also benefits existing methods, we applied it to the final reconstruction residuals of representative unsupervised baselines, retaining only deviations consistent with lower ADC in the observed image than in the reconstruction. This output-level modification left the original models and reconstruction procedures unchanged (Table~\ref{tab:unsupervised_bio_prior}). Compared with their original results in Table~\ref{tab:unsupervised_results}, directional filtering increased Dice from $0.159$ to $0.172$ for DAE, from $0.085$ to $0.104$ for AutoDDPM, and from $0.166$ to $0.173$ for MAD-AD, while pDDPM remained unchanged at the reported precision ($0.121$). Among these modified baselines, MAD-AD achieved the highest Dice ($0.173$) and NSD ($0.226$) and the lowest MASD ($11.16$~mm), whereas DAE achieved the highest AP ($0.680$) and the lowest HD ($25.13$~mm).

Our method with mean filtering achieved a Dice score of $0.283$, significantly exceeding each low-ADC direction filtered baseline (all $p<0.001$, $n=44$). It also achieved a lower MASD ($9.40$~mm) and higher NSD ($0.355$) than all modified baselines. Its HD ($25.45$~mm) was slightly higher than that of DAE, while its AP ($0.505$) exceeded those of AutoDDPM, pDDPM, and MAD-AD but remained below DAE. Thus, our model performance advantage persisted when the same pathological direction was incorporated into baseline outputs.

\section{Discussion}\label{sec:discussion}

\subsection{Overall Detection Performance}

The proposed ADC-only framework demonstrated strong overall performance for unsupervised HIE lesion localization. However, the improvement was not uniform across all evaluation criteria and this behavior reflects the different properties measured by these metrics. AP evaluates the ranking of continuous anomaly scores across lesion and background voxels without selecting a single operating threshold, whereas Dice and the surface metrics evaluate the geometry of the resulting binary segmentation. The comparatively strong Dice and surface performance of our method therefore suggests improved spatial localization, while its lower AP indicates that the continuous anomaly scores remain imperfectly calibrated and that some background voxels still receive scores overlapping those of lesion voxels. Improving cross-subject score calibration and suppressing non-pathological background responses remain important directions for further investigation.

The comparison with supervised ADC-only models also demonstrates the potential of learning from healthy-image structure rather than directly optimizing voxel-wise lesion-mask prediction. Competitive performance in selected spatial
metrics suggests that a learned healthy prior, combined with subject conditioning and pathological directionality, can provide informative localization cues without
voxel-wise lesion supervision in the training objective. This is particularly relevant when available annotations do not adequately represent the diversity of lesion appearances. The findings support healthy-prior modeling
as a complementary route to quantitative HIE lesion assessment.

The lesion-burden analysis indicates that the primary limitation of the proposed method is the detection of subjects with small lesion burdens. Larger HIE abnormalities affect more voxels and are more likely to produce spatially coherent, persistent discrepancies across multiple diffusion noise levels. In contrast, small or low-contrast lesions generate weaker and more localized responses that can be difficult to distinguish from normal anatomical variability, registration differences, and acquisition noise. Mean filtering partially mitigated this limitation by suppressing isolated high-response voxels and reinforcing discrepancies supported by neighboring voxels, resulting in the largest improvements for the small- and medium- burden groups. For subjects with large lesions, however, the anomaly signal was already spatially coherent, leaving less noise for the filter to remove. Additional smoothing could also blur lesion boundaries or extend responses into adjacent healthy tissue, explaining the minimal Dice improvement and the reduction in NSD observed in this group.

The relatively large inter-subject standard deviations further indicate heterogeneous performance across the test cohort. Such variability may reflect differences in lesion burden, anatomical distribution, scan timing, and acquisition characteristics. Future studies should therefore evaluate performance after stratification by lesion burden, injury pattern, scanner field strength, and postnatal age at MRI, in addition to validating the method in independent multi-center cohorts.

\subsection{Intermediate Denoising Discrepancy versus Reconstruction Difference}

Most diffusion-based anomaly-detection methods identify abnormalities by comparing an observed image with its final pseudo-healthy reconstruction. Although intuitive, this formulation compresses the entire reverse diffusion trajectory into a single terminal image. Subtle pathological information expressed during intermediate denoising steps may therefore be attenuated or lost before the final reconstruction is obtained. Moreover, the reconstruction difference reflects not only lesion restoration but also stochastic sampling variation, incomplete lesion removal, and unintended modification of subject-specific healthy anatomy. The latter can generate false-positive responses even when the reconstructed image appears globally plausible.

Our CFG reconstruction experiments illustrate these limitations. Reducing the guidance scale to \(w=0.8\) encouraged stronger restoration toward the learned healthy distribution and increased AP relative to the DDPM baseline, but substantially degraded Dice and the surface-distance metrics. This suggests that weaker subject conditioning increased sensitivity to image differences while also introducing spatial changes unrelated to the true lesion. Conversely, increasing the guidance scale to \(w=7.5\) preserved and strengthened subject-specific information but did not produce consistently accurate anomaly localization. 

Our proposed method instead measures the conditional--unconditional difference directly in the intermediate noise-prediction space. Since both predictions are evaluated from the same noisy image \(\mathbf{x}_t\), the comparison controls for the common forward corruption and isolates the change in the denoising response induced by the subject-specific condition. The unconditional branch represents the response favored by the learned healthy prior, and the conditional branch retains information from the observed subject. Therefore, their difference provides a more direct measure of how the observed anatomy modifies the healthy denoising field than a comparison between two final reconstructed images. The substantially higher Dice obtained using multi noise-level discrepancy accumulation than using either CFG-based reconstruction supports the utility of extracting anomaly evidence before it is compressed into the final reconstruction.

Evaluating each noise level independently provides an additional advantage over following a single reverse chain. In the chain-dependent formulation, each state inherits stochastic variation and reconstruction error from all preceding transitions, and progressive restoration may attenuate subtle pathological signals before they contribute to later discrepancy estimates. In contrast, independently sampling \(\mathbf{x}_t\) from the observed image connects every comparison to the same subject and prevents errors at one noise level from propagating to another. Accumulation across the selected intermediate noise levels emphasizes discrepancies that persist under different corruption strengths rather than relying on a single potentially unstable observation. Intermediate noise levels are particularly useful because they partially suppress the observed lesion while retaining sufficient subject-specific anatomical information. At substantially higher noise levels, the anatomical condition may become weak, whereas near-clean inputs may provide insufficient separation between the two branch responses.

\subsection{Pathology-Informed Directionality and Interpretability}

A central feature of the proposed framework is that it does not treat all conditional--unconditional discrepancies as equally suggestive of pathology. Acute HIE is commonly associated with cytotoxic edema, restricted diffusion, and reduced ADC values \citep{forbes2000neonatal,bobba2023brain,onda2024quantification}. We therefore retain discrepancy components for which the healthy-prior response predicts a higher ADC value than the patient-conditioned response. This constraint provides a clinically motivated direction to the anomaly score and suppresses deviations that are inconsistent with the expected low-ADC appearance of acute HIE. In contrast to an unsigned reconstruction residual, the resulting score can consequently be interpreted as evidence that the observed tissue deviates from the learned healthy distribution in a direction consistent with water diffusion restriction.

The quantitative analysis on the distribution of discrepancy supports this assumption. At the selected noise levels, the expected positive discrepancy occurred more frequently in annotated lesion voxels than in background voxels. Lesion voxels also exhibited a substantially larger positive discrepancy. Consistent with these observations, incorporating the pathology-informed sign constraint improved lesion localization relative to the corresponding unsigned formulation. These findings demonstrate that the selected direction is statistically associated with annotated HIE abnormalities and is more specific than using unsigned discrepancy magnitude alone.

An additional advantage of the proposed formulation is that the pathological prior is incorporated directly into the intermediate diffusion discrepancy, rather than being applied only to a completed reconstruction-derived anomaly map. Most conventional reconstruction-based methods define direction-agnostic anomaly evidence using an absolute or squared residual between the observed image and its pseudo-healthy reconstruction. Although a low-ADC directional constraint can be applied to the signed reconstruction residual, such output-level filtering is performed only after the reconstruction process has mixed pathological deviations with reconstruction blur, anatomical mismatch, and stochastic generation errors. It can remove some directionally inconsistent responses, but it cannot recover pathological evidence that was attenuated during reconstruction or determine whether the final residual was supported consistently across the diffusion trajectory.

The comparative experiments support this distinction. Applying the same low-ADC directional rule as post-processing to conventional anomaly-detection outputs improved their localization performance, indicating that the pathological prior is useful even at the output level. However, these gains were smaller than those obtained when the directional constraint was integrated into the multi noise-level discrepancy formulation. This difference suggests that biological knowledge is more effective when intermediate denoising responses contribute to the anomaly score, rather than being used only to reject candidates after reconstruction.

False-positive responses in regions with low ADC or interhemispheric intensity asymmetry are consistent with
the intended behavior of the directional constraint,
but also highlight its limited pathological specificity.
The constraint selects voxels with deviations in the expected direction of ADC reduction, but directional agreement alone is insufficient to attribute these deviations to HIE-related injury. Incorporating regional anatomical context and spatially adaptive filtering may help distinguish lesion-associated responses from non-pathological variability.

\subsection{Utility and Limitations of Population \(Z\)-Score Refinement}

The population \(Z\)-map analysis demonstrates that voxel-wise normative information can substantially improve lesion localization when an appropriate reference population and accurate spatial correspondence are available. However, this improvement introduces requirements that may limit generalizability in routine clinical settings. Constructing a meaningful ADC \(Z\)-map requires spatially resolved estimates of the healthy population mean and standard deviation. These reference statistics may vary with postmenstrual age, scanner vendor, magnetic field strength, diffusion acquisition protocol, reconstruction procedure, and preprocessing pipeline. This issue is particularly relevant in neonatal imaging because brain structure and diffusion properties change rapidly during early development.

Voxel-wise \(Z\)-maps also depend on accurate alignment between the observed scan and the normative template. Errors in nonlinear registration may generate artificial deviations near tissue boundaries, ventricles, and rapidly developing anatomical structures. Consequently, a \(Z\)-map derived from one cohort may not transfer reliably to a population with different demographics, acquisition characteristics, or spatial normalization quality. Site-specific recalibration or construction of a new normative reference may therefore be required before deployment.

For these reasons, we consider the \(Z\)-map refinement an optional population-reference assisted extension rather than the principal contribution of the proposed framework. The primary ADC-only model does not require population-level voxel-wise means and standard deviations or the precise nonlinear normalization required to construct a normative \(Z\)-map. Although its localization performance is lower than that of the \(Z\)-map assisted configuration, it reduces dependence on cohort-specific reference data and may therefore offer greater portability across clinical populations and acquisition settings. Future work should evaluate whether normative harmonization, age-conditioned reference models, or registration-robust population representations can retain the localization benefit of \(Z\)-scores while improving cross-cohort generalizability.

\subsection{Limitations and Future Work}

Several limitations should be considered when interpreting the present
findings. First, although lesion annotations were not used as voxel-wise
training targets, they were used to exclude lesion-containing slices when constructing the healthy training subset. The proposed training procedure is therefore annotation-free at the voxel-wise training stage but is not
entirely independent of annotations. Moreover, unannotated slices from an HIE-affected subject may still contain subtle or diffuse abnormalities that were not included in the reference mask, potentially contaminating the learned healthy distribution. Future studies should train the framework using an independently acquired cohort of clinically healthy neonates with ADC imaging. This would eliminate the need for lesion mask-based slice selection and permit a more direct evaluation of fully annotation-independent training.

Second, the present evaluation was conducted using a single cohort and a
fixed pre-processing and registration pipeline. Differences in scanner
manufacturer, field strength, acquisition protocol, reconstruction method,
patient age, and clinical population may alter both the ADC distribution and the model response. The population-derived $Z$-map is particularly
sensitive to the population, scanning protocols and the accuracy of nonlinear registration. Multicenter external validation is therefore required to determine whether the learned healthy prior and operating threshold transfer across institutions. 

The current implementation is based on independently processed 2D axial slices. Although this design reduces computational requirements and increases the number of available training samples, it does not explicitly model through-plane anatomical continuity. This limitation may particularly affect small lesions, lesions with weak contrast in an individual slice, and abnormalities extending obliquely or across multiple axial levels. The slice-wise predictions are subsequently assembled into a 3D volume, but this does not recover contextual information that was unavailable to the model during inference. Extending the framework to 2.5D or fully 3D diffusion architectures may improve spatial consistency and the detection of small or fragmented lesions, although these approaches will require substantially greater memory and training data.

The pathology-informed directional prior also assumes that acute HIE injury is associated with reduced ADC relative to healthy tissue. This assumption is clinically motivated but is not universally valid across all imaging times and injury patterns. ADC values evolve after the hypoxic--ischemic event and may pseudo-normalize during the subacute  \citep{vermeulen2008diffusion}. Some injured regions may therefore exhibit weak, absent, or temporally altered ADC reduction despite clinically meaningful tissue damage. Such lesions may not produce a consistently positive conditional--unconditional discrepancy and could be suppressed by the proposed one-sided scoring rule. Future work could incorporate time from injury, gestational age, and treatment timing into the model.

The hard positive-part operation constitutes another limitation. Setting all opposite-direction discrepancies to zero provides a direct implementation of the reduced ADC prior, but it imposes an abrupt boundary at zero. Some lesion voxels may exhibit weakly negative discrepancies or fluctuate around zero across the selected noise levels because of subtle lesion contrast, model uncertainty, or temporal variation in ADC abnormalities. Hard clipping completely removes these responses, potentially reducing sensitivity to small or low-contrast lesions. 

A future extension could replace hard clipping with a soft directional gate, controlled by a parameter $\eta$, that assigns reduced rather than zero weight to near-zero or slightly opposite-direction responses. This would preserve the preference for the expected HIE direction while accounting for uncertainty in the estimated discrepancy. Population-derived ADC $Z$-maps could also be introduced as a continuous spatial weight at this stage, allowing strongly negative $Z$-scores to provide supporting evidence for weak diffusion discrepancies. 

Finally, the continuous anomaly score is not a calibrated lesion probability. Its magnitude may vary with the sampled noise realization, the selected noise levels, the number of accumulated responses, and the diffusion model's uncertainty. Although geometric aggregation reduces the influence of isolated large responses, it does not eliminate stochastic variability or guarantee a common score scale across subjects and cohorts. Future work should quantify repeatability across random seeds and noise realizations, investigate ensembling or deterministic sampling, and estimate voxel-wise uncertainty. 
\section{Conclusion}\label{sec:conclusion}

We proposed OpenMAP-HIE, a pathology-informed score-based diffusion model for unsupervised localization of neonatal HIE abnormalities on ADC images without using lesion annotations as voxel-wise training targets. Rather than relying on a final pseudo-healthy reconstruction, the framework compares and aggregates patient-conditioned and healthy-prior noise predictions across independently evaluated multi noise levels, retaining discrepancies consistent with the expected ADC reduction in acute HIE. The proposed ADC-only method improved lesion overlap and surface-distance
performance over the evaluated unsupervised baselines and achieved Dice performance comparable to supervised ADC-only segmentation models, while population-derived ADC $Z$-map refinement provided further improvements.



 \appendix


\printcredits
\section*{Acknowledgements}
This publication was made possible by grants from the National Institutes of Health R01NS126549. The contents of this paper are solely the responsibility of the authors and do not necessarily represent the official view of NIH.

\section*{Declaration of competing interest}
Kenichi Oishi is a consultant for Corporate‐M and AnatomyWorks. These arrangements are being managed by Johns Hopkins University in accordance with its conflict‐of‐interest policies.
\section*{Declaration of generative AI and AI-assisted technologies
in the manuscript preparation process}

During the preparation of this work, the authors used ChatGPT (OpenAI) to improve the language and readability of the manuscript. After using this tool, the authors reviewed and edited the content as needed and take full responsibility for the content of the
published article.

\section*{Data availability}
The BONBID-HIE dataset used in this study is publicly available
through the BONBID-HIE challenge website
(\url{https://bonbid-hie2023.grand-challenge.org/}),
subject to the dataset's access conditions and terms of use.

\bibliographystyle{cas-model2-names}

\bibliography{cas-refs}

\end{document}